\documentclass[conference]{IEEEtran}
\usepackage{times}

\usepackage[numbers]{natbib}
\usepackage{multicol}
\usepackage{array}
\usepackage{adjustbox}
\usepackage[bookmarks=true]{hyperref}
\usepackage{graphics}
\usepackage[utf8]{inputenc}
\usepackage{booktabs}
\usepackage{multirow}
\usepackage{xspace}
\newcommand{\ie}{i.e.\@\xspace}
\newcommand{\eg}{e.g.\@\xspace}

\usepackage{amsmath}
\usepackage{amssymb}
\usepackage{amsfonts}
\usepackage{mathtools}
\usepackage[font=small]{caption}
\usepackage{wrapfig}
\usepackage{arrayjob}
\usepackage{flushend}
\usepackage{pifont}
\usepackage{bbm}
\usepackage{transparent}
\usepackage{siunitx}
\usepackage{bm}
\usepackage{caption}
\usepackage{makecell}
\usepackage{subcaption}
\usepackage{xcolor}
\usepackage[table]{xcolor}
\usepackage{colortbl}
\usepackage{grffile}
\usepackage{CJKutf8}
\usepackage{blindtext}
\usepackage[capitalise]{cleveref}
\usepackage{graphicx}
\usepackage{svg}
\usepackage{tcolorbox}
\usepackage{listings}
\usepackage{pgfplots}
\pgfplotsset{compat=1.16}
\usetikzlibrary{intersections}
\usepgfplotslibrary{fillbetween}

\def\BibTeX{{\rm B\kern-.05em{\sc i\kern-.025em b}\kern-.08em
    T\kern-.1667em\lower.7ex\hbox{E}\kern-.125emX}}

\DeclareMathAlphabet\mathbfcal{OMS}{cmsy}{b}{n}
\DeclarePairedDelimiterX{\infdivx}[2]{(}{)}{%
  #1\;\delimsize|\delimsize|\;#2%
}

\crefname{figure}{Fig.}{Figs.}
\Crefname{figure}{Fig.}{Figs.}
\crefname{table}{Tab.}{Tabs.}
\Crefname{table}{Tab.}{Tabs.}

\crefname{section}{Sec.}{Secs.}
\Crefname{section}{Sec.}{Secs.}
\crefname{subsection}{Sec.}{Secs.}
\Crefname{subsection}{Sec.}{Secs.}
\crefname{subsubsection}{Sec.}{Secs.}
\Crefname{subsubsection}{Sec.}{Secs.}

\newcommand{\cmark}{\textcolor{green!60!black}{\ding{51}}} 
\newcommand{\xmark}{\textcolor{red!75!black}{\ding{55}}}   
\newcommand{\yes}{\cmark}
\newcommand{\no}{\xmark}

\usepackage{marvosym}
\newcommand{\corrmark}{\textsuperscript{\Letter}}

\definecolor{lightblue}{RGB}{173,216,230} 
\definecolor{lightred}{RGB}{255,182,193}
\definecolor{customblue}{HTML}{ccf2f5}

\definecolor{c_ceiling}{RGB}{220, 45, 45}
\definecolor{c_floor}{RGB}{ 40,160, 40}
\definecolor{c_wall}{RGB}{155,210,225}
\definecolor{c_window}{RGB}{115,155,210}
\definecolor{c_chair}{RGB}{195,200, 75}
\definecolor{c_bed}{RGB}{255,180,110}
\definecolor{c_sofa}{RGB}{140,105,180}
\definecolor{c_table}{RGB}{ 25,110,180}
\definecolor{c_tvs}{RGB}{150,180, 55}
\definecolor{c_furniture}{RGB}{255,140,  0}
\definecolor{c_objects}{RGB}{195,180,220}

\definecolor{backgray}{gray}{0.9}
\definecolor{lightgray}{gray}{0.5}   
\definecolor{darkgray}{gray}{0.25}    

\newcommand{\clight}{\color{lightgray}}
\newcommand{\cdark}{\color{darkgray}}

\newcommand{\csq}[1]{\textcolor{#1}{\rule{2.8mm}{2.8mm}}}
\newcolumntype{C}[1]{>{\centering\arraybackslash}p{#1}}
\newcommand{\modelname}{FreeOcc}

\let\titleold\title
\renewcommand{\title}[1]{\titleold{#1}\newcommand{\thetitle}{#1}}
\def\maketitlesupplementary
   {
   \newpage
       \twocolumn[
        \centering
        \Large
        \vspace{0.5em}\modelname: Training-Free Embodied Open-Vocabulary Occupancy Prediction \\
        Supplementary Material \\
        \vspace{1.0em}
       ] 
   }

\begin{document}

\title{\modelname: Training-Free Embodied Open-Vocabulary Occupancy Prediction}



\author{
Zeyu Jiang$^{* 1}$ \quad
Changqing Zhou$^{* 1}$ \quad 
Xingxing Zuo$^{2}$ \quad 
Changhao Chen$^{1}$\corrmark \quad \\

$^1$The Hong Kong University of Science and Technology (Guangzhou) \quad
$^2$MBZUAI \quad 
}

\twocolumn[{%
\renewcommand\twocolumn[1][]{#1}%
\maketitle

\begin{center}
\centering
\captionsetup{type=figure}
\vspace{-5mm}
\includegraphics[page=1, width=\textwidth]{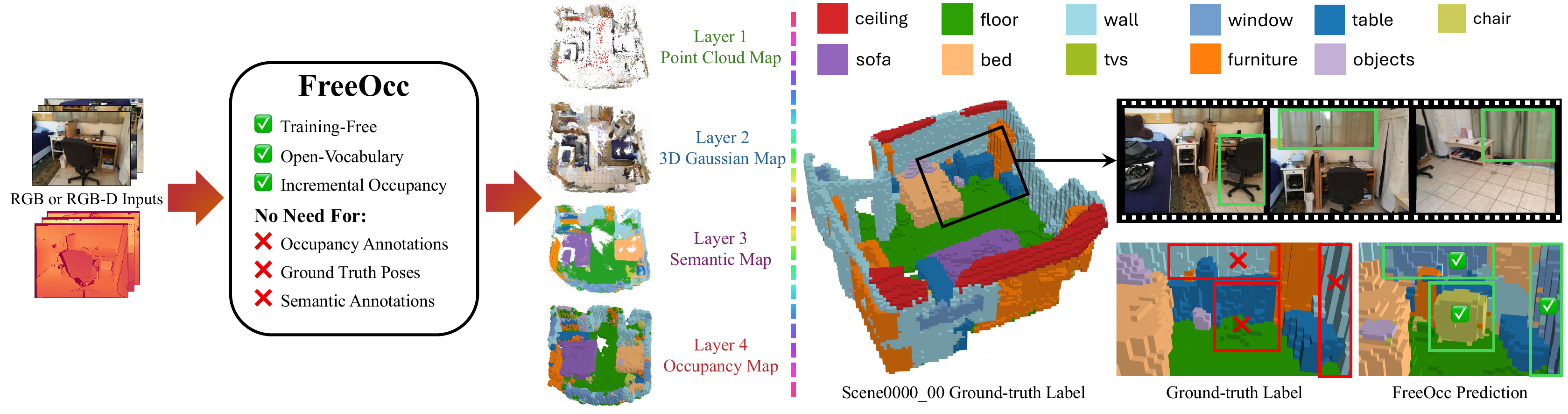} 
\vspace{-5mm}
\caption{\modelname{} is a training-free paradigm for open-vocabulary occupancy prediction. It eliminates the need for occupancy, pose, and semantic annotations, and incrementally constructs four-layer maps using only monocular or RGB-D image sequences. The right panel illustrates the benefit of open-vocabulary reasoning on EmbodiedOcc-ScanNet: the \textcolor{green}{green boxes} corresponding to “window” and “chair” are correctly identified and localized by \modelname{}, whereas the ground-truth occupancy labels (\textcolor{red}{red boxes}) coarsely classify them as “wall” and “floor,” respectively, despite clear visual evidence.}
\label{fig: teaser}
\end{center}
}]

\begingroup
\renewcommand{\thefootnote}{\fnsymbol{footnote}}
\footnotetext[1]{Equal contribution. \quad \Letter\ Corresponding author.}
\endgroup

\begin{abstract}
Existing learning-based occupancy prediction methods rely on large-scale 3D annotations and generalize poorly across environments. We present \modelname{}, a training-free framework for open-vocabulary occupancy prediction from monocular or RGB-D sequences. Unlike prior approaches that require voxel-level supervision and ground-truth camera poses, \modelname{} operates without 3D annotations, pose ground truth, or any learning stage.
\modelname{} incrementally builds a globally consistent occupancy map via a four-layer pipeline: a SLAM backbone estimates poses and sparse geometry; a geometrically consistent Gaussian update constructs dense 3D Gaussian maps; open-vocabulary semantics from off-the-shelf vision–language models are associated with Gaussian primitives; and a probabilistic Gaussian-to-occupancy projection produces dense voxel occupancy.
Despite being entirely training-free and pose-agnostic, \modelname{} achieves over $2\times$ improvements in IoU and mIoU on EmbodiedOcc-ScanNet compared to prior self-supervised methods. We further introduce ReplicaOcc, a benchmark for indoor open-vocabulary occupancy prediction, and show that \modelname{} transfers zero-shot to novel environments, substantially outperforming both supervised and self-supervised baselines. Project page: \url{https://the-masses.github.io/freeocc-web/}.
\end{abstract}

\IEEEpeerreviewmaketitle

\section{Introduction}
\label{sec:intro}

The ability to construct and understand the environment from egocentric observations is fundamental to embodied lifelong autonomy. Beyond geometric reconstruction, robots require scene representations that capture dense structure, semantic completeness, and global spatial consistency to support navigation and interaction in open environments~\cite{deng2026best3dscenerepresentation}. Point clouds arise naturally in depth sensing, structure-from-motion~\cite{ullman1979interpretation}, and SLAM~\cite{cadena2017past, lv2021clins}, and have long served as a core representation for robotic perception~\cite{chen1989representation}. However, their unstructured nature and irregular sampling limit their effectiveness for downstream reasoning.
3D Gaussian Splatting (3DGS)~\cite{kerbl3Dgaussians} addresses these limitations by augmenting each 3D sample with anisotropic extent and opacity, providing a compact continuous representation suitable for differentiable rendering.
Despite its advantages, 3DGS is typically optimized under photometric supervision, often resulting in inaccurate geometry, depth distortions, and view-inconsistent artifacts~\cite{poggi2024selfevolvingdepthsupervised3d,ververas2024sags,pan2025pings}. Moreover, geometric boundaries are commonly extracted via heuristic density thresholds, yielding ambiguous or inconsistent object extents~\cite{fei20243d}.

In contrast, occupancy maps discretize space into free, occupied, and unknown regions, providing explicit geometric boundaries and supporting incremental updates~\cite{hornung13auro,song2016ssc}. Since these boundaries are directly used for collision checking and motion planning, their accuracy is critical for safety and task performance~\cite{frieder1985back}. Motivated by the complementary strengths of Gaussians and occupancy, recent works combine 3D Gaussian primitives with voxelized occupancy representations~\cite{GaussTR,gaussianformer,gaussianformer2}, leading to the embodied occupancy prediction task~\cite{embodiedocc}, which estimates semantic occupancy volumes from Gaussians built from egocentric observations.

Embodied occupancy prediction has advanced rapidly, achieving strong geometric and semantic accuracy in indoor environments and, in some cases, real-time inference~\cite{embodiedocc,embodiedocc++,roboocc,li2025enhancing}. However, most existing methods rely on fully supervised voxel-level annotations and assume accurate camera poses at inference~\cite{ISO,scannet}. Such supervision is expensive, requiring large-scale reconstruction and labeling, and methods trained in this regime often generalize poorly beyond the training distribution. While recent self-supervised approaches reduce annotation requirements and introduce open-vocabulary semantics via vision–language models~\cite{GaussTR,gan2024gaussianocc,clip,roman}, learning-based methods still depend on accurate poses during training and inference and tend to overfit to specific scenes, viewpoints, or sensor configurations, leading to significant performance degradation in novel environments.

To address these limitations, we propose \modelname, the first training-free framework for open-vocabulary occupancy prediction, which incrementally constructs a globally consistent occupancy map through a four-layer mapping pipeline.
\textbf{Layer 1:} A SLAM backbone processes monocular or RGB-D image sequences to estimate camera poses and build sparse point cloud maps.
\textbf{Layer 2:} We construct dense 3D Gaussian maps using SLAM-guided point initialization and a geometrically consistent Gaussian update strategy, ensuring structural fidelity and long-term global consistency.
\textbf{Layer 3:} Open-vocabulary semantic features extracted from pre-trained vision–language models (VLMs) are incrementally associated with Gaussian primitives, enabling language-based querying without voxel-level supervision.
\textbf{Layer 4:} A probabilistic Gaussian-to-occupancy projection aggregates geometric and semantic evidence into a discrete voxel grid, yielding a dense occupancy map with open-vocabulary semantics.
On the EmbodiedOcc-ScanNet benchmark, \modelname{} achieves over \textbf{2$\times$} improvements in both IoU and mIoU compared to self-supervised methods. We further introduce ReplicaOcc, a new benchmark for evaluating generalization in indoor open-vocabulary occupancy prediction, on which \modelname{} demonstrates strong zero-shot generalization, significantly outperforming both supervised and self-supervised learning-based baselines across all metrics.

In summary, we make the following contributions:
\begin{itemize}
    \item We propose \textbf{\modelname}, a training-free framework for open-vocabulary occupancy prediction that addresses the poor generalization of existing occupancy prediction methods caused by dataset-specific training and closed-set supervision.
    \item We identify the geometric ambiguity arising from decoupled 3DGS-SLAM optimization and introduce a novel globally consistent Gaussian update strategy, where geometrically anchored updates produce more spatially consistent 3D scene representations for occupancy mapping.
    \item We present \textbf{ReplicaOcc}, a new benchmark for evaluating generalization in open-vocabulary occupancy prediction. Extensive experiments demonstrate that \modelname\ significantly outperforms prior self-supervised methods and substantially improves generalization compared to learning-based approaches.
\end{itemize}

\section{Related Work}
\label{sec:related}
\subsection{Fully Supervised Occupancy Prediction}

Vision-based occupancy prediction has been widely studied, initially in outdoor autonomous driving benchmarks~\cite{sparseocc,gaussianformer,gaussianformer2,voxformer} and more recently in indoor and embodied environments~\cite{monoscene,ndcscene,sscnet,ISO,embodiedocc,embodiedocc++,zhou2026generalizing,zhou2026monocular}. Fully supervised methods lift 2D image features into 3D representations using depth distributions, ray-based projection, or volumetric aggregation~\cite{monoscene,ndcscene,ISO,fbocc,lss}. Transformer-based volumetric models capture long-range dependencies~\cite{occupancypoints}, while sparsity-aware designs prune empty regions and process sparse voxels via sparse convolutions~\cite{sparseocc} or efficient Transformers~\cite{voxformer,octreeocc}.
Despite strong performance, these approaches rely on dense voxel-level annotations that are expensive to obtain and difficult to scale. Moreover, full supervision often leads to limited generalization to novel scenes or sensor configurations.

\subsection{Weakly Supervised Occupancy Prediction}

Weakly supervised methods reduce annotation costs by learning occupancy from indirect supervision such as 2D segmentation, sparse LiDAR, or pseudo-labels~\cite{lange2024self}. Several approaches optimize 3D occupancy using 2D-only supervision via differentiable rendering or image-space distillation~\cite{pan2024renderocc,boeder2025gaussianflowoccsparseweakly,li2025ago}. Others construct approximate 3D supervision by aggregating sparse LiDAR points~\cite{veon,gao2025loc,pop3d,li2025ago} or enforce multi-view consistency through image-plane reprojection~\cite{selfocc,occnerf,liu2024gausstrfoundationmodelaligned,boeder2024langocc,gan2024gaussianocc}.
While weak supervision alleviates labeling requirements, most methods assume accurate camera poses, operate in fixed domains, and perform offline inference, limiting their applicability to embodied agents that require online, incremental mapping~\cite{embodiedocc}.

\subsection{3D Gaussian Splatting SLAM}
3D Gaussian Splatting SLAM (3DGS-SLAM) jointly estimates camera poses and optimizes continuous Gaussian maps, and has recently attracted significant attention~\cite{Yan_2024_CVPR,Huang_2024_CVPR,Keetha_2024_CVPR,Matsuki:Murai:etal:CVPR2024,yugay2023gaussianslam,lang2025gaussian,lang2025gaussianlic2,li2025pg}. Its continuous, differentiable, and compact representation~\cite{kerbl3Dgaussians} makes it well-suited as a geometric carrier for occupancy prediction~\cite{gaussianformer}. Extensions to semantic and open-set 3DGS-SLAM further enable semantic self-supervision and open-vocabulary reasoning~\cite{li2024sgs,Ji_2024,li2024gs3lam,yang2025opengsslamopensetdensesemantic}.
However, existing 3DGS-SLAM methods primarily optimize photometric objectives for view synthesis, often resulting in spatial inconsistencies that limit voxel-level completion~\cite{Keetha_2024_CVPR, 10.1007/978-3-031-72764-1_11, Huang_2024_CVPR}. In contrast, our work introduces a geometrically consistent 3DGS update strategy and leverages it as a prior for training-free, open-vocabulary occupancy prediction, bridging continuous surface mapping and discrete volumetric reasoning.

\section{Problem Statement}
\label{sec:preliminaries}

\begin{table}[t]
\centering
\caption{Comparison of supervision, inference inputs, and outputs across methods. \textbf{O}: human-labeled occupancy; \textbf{S}: human-labeled semantics; \textbf{P}: human-labeled poses; \textbf{D}: depth; \textbf{R}: RGB images. Outputs include \textbf{OV} (open-vocabulary semantics) and \textbf{OCC} (occupancy prediction).}
\label{tab:setting_compact_singlecol}
\footnotesize
\renewcommand{\arraystretch}{0.9}
\setlength{\tabcolsep}{4pt}
\begin{tabular}{lccc}
\toprule
\textbf{Method} & \textbf{Train} & \textbf{Infer} & \textbf{Output} \\
\midrule
\rowcolor{backgray}
\multicolumn{4}{l}{\textit{Fully supervised}} \\
EmbodiedOcc~\cite{embodiedocc}   & O,S,P,D,R & P,R & OCC \\
EmbodiedOcc++~\cite{embodiedocc++} & O,S,P,D,R & P,R & OCC \\
RoboOcc~\cite{roboocc}       & O,S,P,D,R & P,R & OCC \\
\midrule
\rowcolor{backgray}
\multicolumn{4}{l}{\textit{Self-supervised}} \\
GaussTR~\cite{GaussTR}       & P,R & P,R & OV,OCC \\
GaussianOCC~\cite{gan2024gaussianocc}   & P,R & P,R & OV,OCC \\
\midrule
\rowcolor{backgray}
\multicolumn{4}{l}{\textit{Training-free}} \\
FreeOcc (mono)& --  & R   & OV,OCC \\
FreeOcc (rgbd)& --  & D,R & OV,OCC \\
\bottomrule
\end{tabular}
\vspace{-15pt}
\end{table}

\noindent\textbf{Task Overview.}
\modelname~addresses the problem of \emph{embodied} semantic occupancy prediction~\cite{embodiedocc,embodiedocc++,roboocc}. In contrast to monocular scene completion methods that infer a 3D occupancy map from a single RGB image~\cite{sscnet,wang2023semantic,ISO}, the embodied setting requires a robot to \emph{incrementally} construct a \emph{globally consistent} semantic occupancy map from egocentric observations while actively exploring the environment.

Formally, given a stream of RGB observations $\mathcal{I}_{1:T}={\mathcal{I}_1,\mathcal{I}_2,\dots,\mathcal{I}_T}$, our goal is to estimate a global 3D semantic occupancy field $\mathcal{O}_T \in \mathbb{R}^{X\times Y\times Z\times C}$ in an online manner. Here, $(X,Y,Z)$ denote the spatial resolution of the scene volume, and $C$ is the number of semantic categories. Each voxel encodes both geometric occupancy (occupied or free) and semantic evidence, representing the environment observed up to time $T$.

\noindent\textbf{Key Differences from Prior Work.}
As summarized in Tab.~\ref{tab:setting_compact_singlecol}, existing occupancy prediction approaches can be broadly categorized according to the supervision and prior information required during training and inference:
\begin{itemize}
\item \textit{Fully supervised} methods~\cite{embodiedocc, embodiedocc++, roboocc} rely on dense voxel-level annotations, which are costly to obtain and typically require large-scale 3D reconstruction followed by manual or semi-automatic labeling pipelines.
\item \textit{Self-supervised} methods~\cite{GaussTR, gan2024gaussianocc} reduce dependence on voxel annotations but still assume \emph{known camera poses} during both training and inference. Moreover, as learning-based approaches, both supervised and self-supervised methods often suffer from limited cross-scene generalization, leading to degraded zero-shot performance in previously unseen environments, as demonstrated in~\Cref{sec: occ_evaluation}.
\end{itemize}

In contrast, \textbf{\modelname} is a \textit{training-free} approach that requires neither semantic annotations nor prespecified camera poses and directly predicts embodied occupancy from an incoming video stream. Furthermore, since RGB and depth sequences are commonly available in robotic systems, \modelname~supports two inference modes: monocular RGB and RGB-D, enabling flexible deployment across diverse sensing platforms.

\section{Methodology}

\begin{figure*}[htb!]
\centering
\captionsetup{font=small}
\includegraphics[page=1, width=\linewidth]{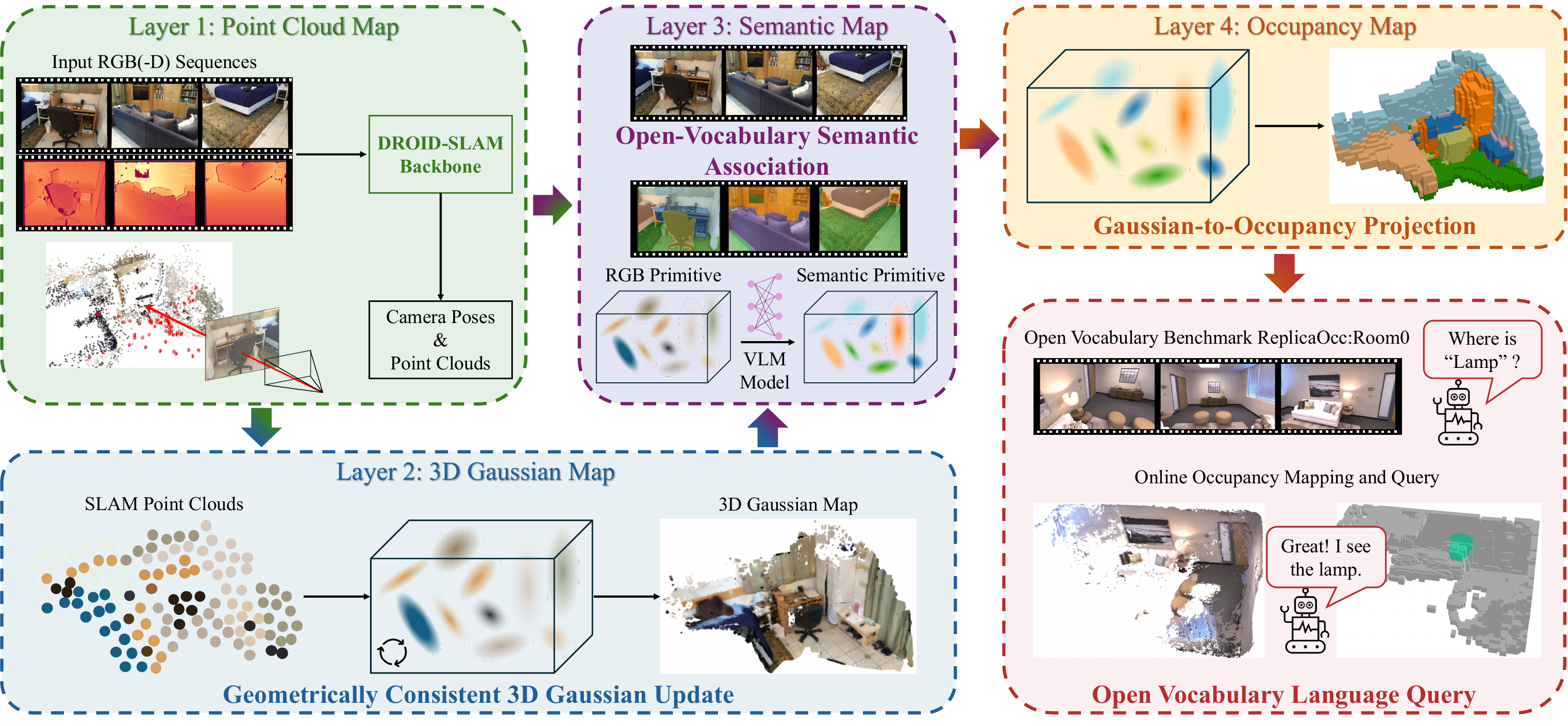} 
\caption{\textbf{Framework Overview of \modelname.} \modelname~incrementally constructs a multi-layer map for online open-vocabulary occupancy prediction.
\textbf{Layer 1:} A SLAM backbone processes monocular or RGB-D image sequences to estimate camera poses and sparse/semi-dense point cloud maps.
\textbf{Layer 2:} Dense 3D Gaussian Splatting (3DGS) maps are constructed via SLAM-guided point initialization and a geometrically consistent Gaussian update strategy.
\textbf{Layer 3:} Open-vocabulary semantic features are associated with Gaussian primitives using a vision–language model, forming a language-embedded 3D Gaussian semantic map.
\textbf{Layer 4:} The semantic Gaussian map is projected into a dense voxel occupancy representation through probabilistic Gaussian-to-occupancy splatting, enabling online open-vocabulary querying and semantic localization in 3D scenes.}
\label{fig: pipeline}
\vspace{-15pt}
\end{figure*}

In this section, we first present an overview of the \modelname~system architecture in \cref{sec: architecture}. We then detail the geometrically consistent 3D Gaussian mapping process in \cref{sec: gs_mapping}. Next, we introduce the open-vocabulary semantic association module in \cref{sec: ov}, which injects language-aligned semantics into the Gaussian map to form language-embedded (LE) Gaussians. Finally, we describe how the LE-Gaussian representation is converted into a volumetric occupancy map in \cref{sec: gs2occ}, enabling open-vocabulary querying with arbitrary text prompts.

\subsection{Overall Architecture of \modelname}
\label{sec: architecture}

The overall architecture of \modelname~is illustrated in \cref{fig: pipeline}. At a high level, \modelname~is an online, training-free system that incrementally constructs an open-vocabulary 3D occupancy map from monocular or RGB-D image streams.
Specifically, \modelname~explicitly maintains multi-level scene representations and continuously refines them as new observations arrive, in a fully streaming fashion. It consists of four tightly coupled components:
\textbf{Layer 1} a SLAM backbone for globally consistent camera pose estimation and geometric reconstruction,
\textbf{Layer 2} a geometrically anchored 3D Gaussian mapping module that lifts SLAM point clouds into a continuous scene representation,
\textbf{Layer 3} an open-vocabulary semantic association module that assigns language-aligned embeddings to Gaussian primitives, and
\textbf{Layer 4} a Gaussian-to-occupancy projection module that converts the Gaussian representation into a volumetric occupancy field.
We describe each component in detail below.

\noindent\textbf{Geometric Correspondence using SLAM Backbone.}\label{sec: SLAM}
\modelname~first processes incoming observations using a SLAM backbone to estimate camera poses and sparse 3D geometry. In principle, the proposed framework is compatible with arbitrary SLAM systems. In this work, we adopt DROID-SLAM~\cite{teed2021droid} due to its strong global geometric consistency and robustness under monocular input. Unlike feed-forward, model-based SLAM approaches such as MASt3R-SLAM~\cite{murai2024_mast3rslam} or VGGT-SLAM~\cite{maggio2025vggt-slam}, DROID-SLAM does not rely on explicit 3D supervision from structure-from-motion (SfM) pipelines~\cite{schoenberger2016sfm} during training of its optical flow network~\cite{624b677c512440f9816f866607d16018}. By jointly optimizing over long temporal windows, DROID-SLAM produces globally consistent camera poses  $\mathcal{T}_{1:T} = \{ \mathbf{T}_1, \dots, \mathbf{T}_T \}$ and an accumulated set of 3D points
$\mathcal{P}_{1:T} = \{ \mathbf{p}_i \in \mathbb{R}^3 \}_{i=1}^{N_T}$ which together provide a stable spatial reference for downstream mapping. This global consistency allows all subsequent modules to operate in a unified coordinate frame and is critical for mitigating geometric drift in long-horizon mapping.

\noindent\textbf{Geometrically Consistent 3D Gaussian Construction.}
Given globally consistent camera poses $\mathcal{T}_{1:T}$ and 3D points $\mathcal{P}_{1:T}$ from SLAM, \modelname~incrementally maintains a set of 3D Gaussian primitives $\mathcal{G} = \{ G_i \}_{i=1}^{N_T}$.We update the Gaussians while preserving multi-view geometric consistency rather than optimizing for novel-view synthesis as in prior 3DGS-SLAM~\cite{Homeyer_2025_ICCV}. The resulting Gaussian map bridges sparse SLAM geometry and dense volumetric occupancy reasoning.

\noindent\textbf{Open-Vocabulary Semantic Association.}
We extract open-vocabulary semantics from 2D observations using pre-trained vision--language models~\cite{clip,trident} and fuse them into Gaussian primitives via geometric correspondence. The aggregated semantics are propagated to the occupancy grid for language-driven querying without voxel-level supervision.

\noindent\textbf{Gaussian-to-Occupancy Projection.}
Finally, the maintained Gaussian set $\mathcal{G}$ is projected into a discrete occupancy field
$\mathcal{O} \in \mathbb{R}^{X \times Y \times Z \times C}$
by aggregating the probabilistic spatial support of nearby Gaussian primitives.

\subsection{Geometrically Consistent 3D Gaussian Construction}
\label{sec: gs_mapping}

\noindent\textbf{Geometric Ambiguity Problem.} Recent 3DGS-SLAM approaches grounded in point-based SLAM systems—such as Photo-SLAM~\cite{Huang_2024_CVPR} built upon ORB-SLAM~\cite{murORB2}, and DROID-Splat~\cite{teed2021droid} built upon DROID-SLAM~\cite{teed2021droid}—naturally inherit the localization accuracy and efficiency of classical SLAM pipelines. However, these methods share a fundamental property: the Gaussian scene representation is optimized independently of the point-based SLAM map. In such decoupled systems, Gaussian parameters are primarily updated to maintain rendering consistency, while the SLAM backend enforces geometric consistency~\cite{11127380}.

Following standard 3DGS formulations, we represent the scene as a set of language-embedded Gaussian primitives,
each parameterized as $G_i=(\boldsymbol{\mu}_i,\mathbf{s}_i,\mathbf{r}_i,o_i,\mathbf{c}_i,\mathbf{f}_i)$,
where $\boldsymbol{\mu}_i\in\mathbb{R}^3$ is the 3D mean, $\mathbf{s}_i\in\mathbb{R}_+^3$ the anisotropic scale,
$\mathbf{r}_i$ the rotation, $o_i$ the opacity, $\mathbf{c}_i$ the color, and $\mathbf{f}_i$ the
language-aligned open-vocabulary feature.
Given camera intrinsics $K_{1:T}$ and globally consistent poses $\mathcal{T}_{1:T}$, we follow~\cite{gaussiansplatting}
to render the Gaussian set $\mathcal{G}$ into each view:
\begin{equation}
\label{eq:render}
(\hat{I}_t,\hat{D}_t) \;=\; F(\mathcal{G}, K_t, \mathcal{T}_t), \qquad t=1,\ldots,T,
\end{equation}
where $F$ denotes the differentiable Gaussian rendering operator and $\hat{I}_t,\hat{D}_t$ represent render image and depth map. Let $\theta$ collect all Gaussian parameters.
The rendered outputs are compared against observations $\{(I_t,D_t)\}_{t=1}^{T}$, and typical 3DGS-based mapping solves
\begin{equation}
\label{equa:1}
\min_{\theta}\; \sum_{t=1}^{T}\Big(\big\|\hat{I}_t - I_t\big\|_2^2 + \beta\,\big\|\hat{D}_t - D_t\big\|_2^2\Big),
\end{equation}
where $\beta$ balances the RGB and depth terms.

Let $\theta^\star$ denote a solution to Eq.~\eqref{equa:1}. Linearizing the rendering operator around $\theta^\star$ yields
$F(\theta^\star + \delta \theta) \approx F(\theta^\star) + J\,\delta \theta$,
where $J = \left.\frac{\partial F}{\partial \theta}\right|_{\theta^\star}$.
For unobservable or weakly observable directions, there exist non-zero perturbations $\delta \theta \neq 0$ such that $J\,\delta \theta = 0$.
Consequently, the solution to Eq.~\eqref{equa:1} is not isolated, but instead lies on a (near-)manifold in parameter space.

This ambiguity can be further illustrated by considering a single pixel ray $\mathbf{u}$. Under volumetric alpha compositing~\cite{10.1145/3503250}, the rendered color and depth along $\mathbf{u}$ can be written as
\begin{equation}
\label{eq:ray_comp}
\hat{I}(\mathbf{u}) \;=\; \sum_k w_k(\theta;\mathbf{u})\, \mathbf{c}_k,
\qquad
\hat{D}(\mathbf{u}) \;=\; \sum_k w_k(\theta;\mathbf{u})\, z_k,
\end{equation}
where $w_k(\theta;\mathbf{u}) \ge 0$ are ray-dependent compositing weights, and $z_k$ denotes the depth of the $k$-th contributing Gaussian along the ray. Eq.~\eqref{eq:ray_comp} constrains only the first-order moments along each ray: multiple distinct depth--weight configurations $\{(w_k,z_k)\}$ can yield identical $(\hat{I}(\mathbf{u}),\hat{D}(\mathbf{u}))$. Consequently, even with depth supervision, multiple distinct Gaussian configurations $\mathcal{G}$ (equivalently, parameter sets $\theta$) may explain the same observations $\{(I_t,D_t)\}_{t=1}^{T}$. Therefore, minimizing the rendering loss in Eq.~\eqref{equa:1} does not guarantee a unique or globally consistent 3D geometry, and unconstrained Gaussian updates can gradually erode the geometric consistency provided by the SLAM backend.

\noindent\textbf{Geometrically Anchored Gaussian Updates.}
To address the above ambiguity, we propose a geometrically consistent 3D Gaussian update strategy with geometry-aware initialization. We parameterize each Gaussian ellipsoid by an anisotropic scale vector $\mathbf{s}$, which controls its principal axis lengths.

For a pixel $\mathbf{u}$ in frame $t$, we compute the normalized ray direction $\mathbf{d}_{t,\mathbf{u}}\in\mathbb{R}^3$ from intrinsic $K_t$.
We define a local rotation $R_{t,\mathbf{u}}$ such that its local $+Z$ axis aligns with $\mathbf{d}_{t,\mathbf{u}}$.
We initialize a ray-aligned anisotropic scale as
\begin{equation}
\mathbf{s}_{t,u} = (s_\perp,\, s_\perp,\, s_\parallel),
\qquad
s_\parallel = \gamma\, s_\perp ,
\end{equation}
where $s_\perp$ and $s_\parallel$ denote the Gaussian extents perpendicular and parallel to the viewing ray, respectively, and $\gamma$ is a user-controlled elongation ratio. This initialization models each Gaussian as a thin ellipsoid aligned with the sensor ray, providing a geometry-aware prior that reduces ambiguity during optimization.
Given SLAM-estimated camera poses
$\mathcal{T}_t$ and 3D point positions $\mathcal{P}_t$,  Gaussian centers $\boldsymbol{\mu}$ are fixed to $\mathcal{P}_t$, and the optimization problem is formulated as
\begin{equation}
\begin{aligned}
\min_{\theta}\; \sum_{t=1}^{T}\Big(\big\|\hat{I}_t - I_t\big\|_2^2 + \beta\,\big\|\hat{D}_t - D_t\big\|_2^2\Big), \,\,
\text{s.t.}\;&
\boldsymbol{\mu}_t=\mathcal{P}_t
\end{aligned}
\end{equation}

\subsection{Open-vocabulary Semantic Association}
\label{sec: ov}

In contrast to conventional occupancy estimation pipelines that predict \emph{closed-set} semantic classes, a key goal of \modelname~is to support \emph{open-vocabulary} 3D querying without committing to a fixed label space.
To this end, we leverage a pre-trained open-vocabulary segmentation model~\cite{trident}.
Such OV segmentation models produce a \emph{language-aligned} embedding for each pixel, and enable open-vocabulary recognition by computing the similarity between the per-pixel embedding and a text embedding extracted by a language encoder (\eg, CLIP~\cite{clip}), thereby localizing the region corresponding to an arbitrary textual prompt.
Building on this capability, we associate each 3D Gaussian $G_i$ with a language-aligned embedding and construct \emph{language-embedded (LE) Gaussians}~\cite{zhou2024feature,shi2024language}.

Specifically, given an input image $\mathcal{I}_t$, we extract dense per-pixel embeddings $\mathbf{z}_t(\mathbf{u})\in\mathbb{R}^{D}$ at pixel coordinate $\mathbf{u}$ using the OV segmentation model~\cite{trident}.
We then lift these pixel-wise embeddings into 3D using the depth estimated by our SLAM module.
For each lifted 3D point, we identify its associated \emph{geometrically anchored} Gaussian in the current Gaussian map and attach the corresponding language-aligned feature to it.
The resulting per-Gaussian language features are stored in the Gaussian map, and can be efficiently queried by arbitrary text prompts during the subsequent Gaussian-to-occupancy projection.

\subsection{Gaussian-to-Occupancy Projection}
\label{sec: gs2occ}
To convert the continuous LE-Gaussian scene representation into a dense volumetric occupancy map, we follow the Gaussian-to-occupancy projection paradigm of GaussianFormer2~\cite{gaussianformer2}.
Specifically, we first determine voxel occupancy from the spatial geometry of 3D Gaussian primitives, \ie, their locations and extents in the scene, to estimate whether each voxel is occupied.
Different from GaussianFormer2~\cite{gaussianformer2}, which targets \emph{closed-set} semantics by aggregating per-class probabilities from Gaussians, we instead construct a \emph{language-embedded} occupancy representation from our LE-Gaussians.
Building on the geometric occupancy, we propagate the per-Gaussian language-aligned features into the voxel grid, yielding a \emph{language-embedded occupancy} (LE-occupancy) map that supports open-vocabulary querying.

Concretely, the procedure is as follows.
For a query 3D location $\mathbf{x}$, we retrieve its neighboring LE-Gaussian primitives
\(
\mathcal{H}(\mathbf{x})=\{G_k\}_{k=1}^{P(\mathbf{x})},
\)
where \(P(\mathbf{x})=|\mathcal{H}(\mathbf{x})|\),
and induces the covariance
\(
\boldsymbol{\Sigma}_k
=R(\mathbf{r}_k)\,\mathrm{diag}(\mathbf{s}_k^2)\,R(\mathbf{r}_k)^\top,
\)
then each neighbor primitive contributes a spatial support
\begin{equation}
\alpha_k(\mathbf{x})
=
\exp\!\left(
-\frac{1}{2}(\mathbf{x}-\boldsymbol{\mu}_k)^\top \boldsymbol{\Sigma}_k^{-1}(\mathbf{x}-\boldsymbol{\mu}_k)
\right),
\,\, G_k \in \mathcal{H}(\mathbf{x}),
\end{equation}
and we compose them using probabilistic exclusion:
\begin{equation}
\alpha(\mathbf{x}) = 1 - \prod_{G_k \in \mathcal{H}(\mathbf{x})}\big(1-\alpha_k(\mathbf{x})\big).
\end{equation}
To propagate semantics, we compute the posterior responsibility under a local Gaussian mixture model,
\begin{equation}
p(G_k\mid \mathbf{x}) =
\frac{p(\mathbf{x}\mid G_k)\,\pi_k}{\sum_{G_j \in \mathcal{H}(\mathbf{x})} p(\mathbf{x}\mid G_j)\,\pi_j},
\end{equation}
where $p(\mathbf{x}\mid G_k)=\mathcal{N}\!\left(\mathbf{x};\boldsymbol{\mu}_k,\boldsymbol{\Sigma}_k\right)$ and $\pi_k$ is the mixture weight (we set $\pi_k=o_k$, \ie, opacity).
We propagate the per-primitive language features to the voxel location by posterior expectation:
\begin{equation}
\mathbf{f}(\mathbf{x})=\sum_{G_k \in \mathcal{H}(\mathbf{x})} p(G_k\mid \mathbf{x})\,\mathbf{f}_k,
\qquad
\hat{\mathbf{f}}(\mathbf{x})=\frac{\mathbf{f}(\mathbf{x})}{\|\mathbf{f}(\mathbf{x})\|_2}.
\end{equation}
Given a queried category set $\mathcal{C}$, we obtain the corresponding text embeddings $\{\mathbf{t}_c\}_{c\in\mathcal{C}}$ using a text encoder~\cite{clip} and compute the voxel--text similarity
\begin{equation}
\hat{\mathbf{t}}_c=\frac{\mathbf{t}_c}{\|\mathbf{t}_c\|_2},
\qquad
s(\mathbf{x},c) \;=\; \hat{\mathbf{f}}(\mathbf{x})^\top \hat{\mathbf{t}}_c.
\end{equation}

This yields an open-vocabulary semantic score at $\mathbf{x}$.
We output $\alpha(\mathbf{x})$ and $s(\mathbf{x},c)$ as the volumetric occupancy probability and open-vocabulary semantic score, respectively. In practice, semantics are reported only for occupied voxels.
 
\begin{table*}[t]
\centering
\caption{Performance comparison on EmbodiedOcc-ScanNet. Label requirements are reported by task type: \textbf{Geo.} indicates required geometric supervision, and \textbf{Sem.} indicates semantic supervision. We report IoU and per-class mIoU.}
\label{tab:main_embodiedocc_scannet}
\scriptsize
\setlength{\tabcolsep}{4.2pt}
\renewcommand{\arraystretch}{0.9}
\begin{adjustbox}{max width=\textwidth}
\begin{tabular}{
l
C{1.10cm} C{0.85cm}
c
*{11}{c}
c
}
\toprule
\multirow{2}{*}{\textbf{Method}} &
\multicolumn{2}{c}{\textbf{Annotation}} &
\multirow{2}{*}{\textbf{IoU}} &
\textbf{ceiling} & \textbf{floor} & \textbf{wall} & \textbf{window} &
\textbf{chair} & \textbf{bed} & \textbf{sofa} & \textbf{table} &
\textbf{tvs} & \textbf{furniture} & \textbf{objects} &
\multirow{2}{*}{\textbf{mIoU}} \\
\cmidrule(lr){2-3}
& \textbf{Geo.} & \textbf{Sem.} &
& \csq{c_ceiling} & \csq{c_floor} & \csq{c_wall} & \csq{c_window} &
  \csq{c_chair} & \csq{c_bed} & \csq{c_sofa} & \csq{c_table} &
  \csq{c_tvs} & \csq{c_furniture} & \csq{c_objects} & \\
\midrule

\multicolumn{15}{l}{\clight\textit{Fully supervised learning}} \\
\midrule
\clight TPVFormer~\cite{Triformer}
& Occupancy & \yes
& \clight 35.88 & \clight 1.62 & \clight 30.54 & \clight 12.03 & \clight 13.22
& \clight 35.47 & \clight 51.39 & \clight 49.79 & \clight 25.63
& \clight 3.6  & \clight 43.15 & \clight 16.23 & \clight 25.70 \\

\clight SurroundOcc~\cite{surroundocc}
& Occupancy & \yes
& \clight 37.04 & \clight 12.7 & \clight 31.8  & \clight 22.5  & \clight 22
& \clight 29.9  & \clight 44.7  & \clight 36.5  & \clight 24.6
& \clight 11.5 & \clight 34.4  & \clight 18.2  & \clight 26.27 \\

\clight GaussianFormer~\cite{gaussianformer}
& Occupancy & \yes
& \clight 38.02 & \clight 17   & \clight 33.6  & \clight 21.5  & \clight 21.7
& \clight 29.4  & \clight 47.8  & \clight 37.1  & \clight 24.3
& \clight 15.5 & \clight 36.2  & \clight 16.8  & \clight 27.36 \\

\clight EmbodiedOcc~\cite{embodiedocc}
& Occupancy & \yes
& \clight 51.52 & \clight 22.7 & \clight \textbf{44.6}  & \clight 37.4  & \clight 38
& \clight 50.1  & \clight 56.7  & \clight 59.7  & \clight 35.4
& \clight 38.4 & \clight 52    & \clight 32.9  & \clight 42.53 \\

\clight EmbodiedOcc++~\cite{embodiedocc++}
& Occupancy & \yes
& \clight 52.2  & \clight \textbf{27.9} & \clight 43.9  & \clight 38.7  & \clight \textbf{40.6}
& \clight 49    & \clight \textbf{57.9}  & \clight 59.2  & \clight \textbf{36.8}
& \clight 37.8 & \clight 53.5  & \clight 34.1  & \clight 43.60 \\

\clight RoboOcc~\cite{roboocc}
& Occupancy & \yes
& \clight \textbf{53.3}  & \clight 21.94& \clight 44.57 & \clight \textbf{39.54} & \clight 38.48
& \clight \textbf{51.28} & \clight 57.04 & \clight \textbf{63.09} & \clight 36.7
& \clight \textbf{43.05} & \clight \textbf{54.42} & \clight \textbf{34.38} & \clight \textbf{44.05} \\

\midrule
\multicolumn{15}{l}{\cdark\textit{Self-supervised learning}} \\
\midrule

\cdark GaussianOcc~\cite{gan2024gaussianocc}
& Poses & \no
& \cdark 10.17 & \cdark 3.81 & \cdark 5.09 & \cdark 2.53
& \cdark 2.56 & \cdark 3.84 & \cdark 10.26 & \cdark 9.9
& \cdark 5.37 & \cdark 0.50 & \cdark 2.33 & \cdark 1.19 & \cdark 4.34 \\

\cdark GaussTR~\cite{GaussTR}
& Poses & \no
& \cdark 15.63 & \cdark 1.20 & \cdark 7.78 & \cdark 4.29 & \cdark 2.67
& \cdark 4.52 & \cdark 11.27 & \cdark 10.95 & \cdark 5.31
& \cdark 0.90 & \cdark 4.21 & \cdark 1.34 & \cdark 4.95 \\

\midrule
\multicolumn{15}{l}{\textit{Training-free}} \\
\midrule
\textbf{Ours (mono)} & \no & \no & 31.29 & 3.16 & 23.49 & 16.14 & 13.11 & 19.66 & 21.64 & 23.43 & 13.76 & 4.01 & 8.04 & 5.98 & 13.86 \\
\textbf{Ours (rgbd)} & \no & \no & 34.40 & 6.56 & 26.46 & 21.69 & 15.15 & 21.02 & 22.09 & 23.61 & 15.87 & 7.48 & 8.28 & 6.00 & 15.84 \\
\bottomrule
\end{tabular}
\end{adjustbox}
\end{table*}

\section{Experiments}
\label{sec:experiments}
In this section, we seek to answer the following research questions:
\begin{itemize}
    \item \textbf{Q1:} Compared with learning-based occupancy prediction methods, can training-free approaches generalize effectively to unseen datasets? (\cref{sec: occ_evaluation})
    \item \textbf{Q2:} Does our Gaussian update strategy offer advantages over well-known 3DGS-SLAM systems in occupancy prediction? How does it impact the FreeOcc system? (\cref{sec: ablation_combined})
    \item \textbf{Q3:} Can our system support online language-based querying of 3D occupancy? (\cref{sec: qualitative})
\end{itemize}

\subsection{ReplicaOcc Benchmark}
\label{replicaocc}
\noindent\textbf{Task Definition.}
We introduce \emph{ReplicaOcc}, a compact, \emph{test-only} benchmark for evaluating \emph{open-vocabulary semantic occupancy prediction} in indoor embodied environments. Similar to EmbodiedOcc-ScanNet~\cite{embodiedocc}, incrementally fuse observations over time to maintain globally consistent occupancy prediction. Unlike local frustum-based prediction methods~\cite{ISO}, this setting emphasizes long-horizon spatial consistency and semantic completeness, which are essential for embodied agents operating in real environments.

\noindent\textbf{Motivation.}
Most existing occupancy prediction methods~\cite{embodiedocc, embodiedocc++, roboocc} are evaluated exclusively on EmbodiedOcc-ScanNet, which also serves as their primary training source. In many 3D vision tasks (e.g., ACE~\cite{brachmann2023ace} and LoFTR~\cite{sun2021loftr}), training on a single dataset such as ScanNet~\cite{dai2017scannet} is often sufficient to generalize to diverse indoor scenes. This motivates the use of a small, test-only dataset to better assess cross-dataset generalization for occupancy prediction.

\noindent\textbf{Dataset Construction.}
We adopt the Replica~\cite{straub2019replica} sequences release from NICE-SLAM~\cite{Zhu2022CVPR}. Following the protocols of Occ-ScanNet~\cite{ISO} and EmbodiedOcc-ScanNet, each scene is converted into a global occupancy grid with resolution \(\mathit{l}_{x}\times\mathit{l}_{y}\times\mathit{l}_{z}/(0.08m)^3\), where \(\mathit{l}_{x}\times\mathit{l}_{y}\times\mathit{l}_{z}\) denotes the spatial extent of the scene in the world coordinate system. Each voxel is annotated with a binary occupancy label and a semantic category.

\subsection{Experimental Setups}
\label{exp_setup}
\noindent\textbf{Datasets and Metrics.}
We evaluate occupancy prediction performance on three datasets: EmbodiedOcc-ScanNet, ReplicaOcc, and EmbodiedOcc-ScanNet-mini. Since each Occ-ScanNet scene clips only 100 frames, it does not satisfy the sequential requirements of SLAM. Therefore, for corresponding scenes, we use monocular or RGB-D sequences from the original ScanNet dataset as SLAM inputs. Evaluation metrics include IoU and mIoU computed on the global scene occupancy, following the EmbodiedOcc evaluation protocol~\cite{embodiedocc}. As fully supervised methods are trained on only 11 semantic categories, mIoU on ReplicaOcc is reported over the 8 categories shared with EmbodiedOcc-ScanNet.

\noindent\textbf{Coordinate System Alignment.} 
During evaluation, inspired by \emph{EVO}~\cite{grupp2017evo}, SLAM-reconstructed maps are aligned with the occupancy ground-truth coordinate system to resolve global gauge freedom.
For monocular or RGB-D SLAM, we estimate a global $\mathrm{Sim}(3)=\{s,\mathbf{R},\mathbf{t}\}$ or $\mathrm{SE}(3)=\{\mathbf{R},\mathbf{t}\}$ by aligning camera centers.
Given matched SLAM and GT camera poses $\{\mathbf{T}_i^{\text{slam}},\mathbf{T}_i^{\text{gt}}\}$, we extract camera centers $\mathbf{c}_i^{\text{slam}},\mathbf{c}_i^{\text{gt}}\in\mathbb{R}^3$ and solve $\min_{s,\mathbf{R},\mathbf{t}} \sum_i \lVert \mathbf{c}_i^{\text{gt}}-(s\,\mathbf{R}\mathbf{c}_i^{\text{slam}}+\mathbf{t}) \rVert_2^2$ using the closed-form Umeyama solution.
The resulting transform is applied consistently to the reconstructed 3D Gaussian map: Gaussian means are updated as $\mathbf{x}'=s\,\mathbf{R}\mathbf{x}+\mathbf{t}$, Gaussian scales as $\boldsymbol{\sigma}'=s\,\boldsymbol{\sigma}$ (or $\log\boldsymbol{\sigma}'=\log\boldsymbol{\sigma}+\log s$), and Gaussian orientations as $\mathbf{R}_g'=\mathbf{R}\mathbf{R}_g$.
This alignment step ensures IoU and mIoU computation without affecting the intrinsic reconstruction quality.

\begin{table*}[t]
\centering
\caption{Zero-shot generalization results on the ReplicaOcc benchmark. The evaluation protocol and categorization follow those in \Cref{tab:main_embodiedocc_scannet}.}
\label{tab:replica_zeroshot_supervision}
\scriptsize
\setlength{\tabcolsep}{4.2pt}
\renewcommand{\arraystretch}{0.9}

\begin{adjustbox}{max width=\textwidth}
\begin{tabular}{
l
C{1.10cm} C{0.85cm}
c
*{8}{c}
c
}
\toprule
\multirow{2}{*}{\textbf{Method}} &
\multicolumn{2}{c}{\textbf{Annotation}} &
\multirow{2}{*}{\textbf{IoU}} &
\textbf{ceiling} & \textbf{floor} & \textbf{wall} & \textbf{window} &
\textbf{chair} & \textbf{bed} & \textbf{sofa} & \textbf{table} &
\multirow{2}{*}{\textbf{mIoU}} \\
\cmidrule(lr){2-3}
& \textbf{Geo.} & \textbf{Sem.} &
& \csq{c_ceiling} & \csq{c_floor} & \csq{c_wall} & \csq{c_window} &
  \csq{c_chair} & \csq{c_bed} & \csq{c_sofa} & \csq{c_table} &
  \multicolumn{1}{c}{} \\
\midrule

\multicolumn{12}{l}{\clight\textit{Fully supervised learning}} \\
\midrule
\clight EmbodiedOcc~\cite{embodiedocc}
& Occupancy & \yes
& \clight 22.91 & \clight 0.00 & \clight 0.00 & \clight 0.00 & \clight 0.01
& \clight 0.00 & \clight 0.00 & \clight 0.01 & \clight 0.01 & \clight 0.00 \\


\midrule
\multicolumn{12}{l}{\cdark\textit{Self-supervised learning}} \\
\midrule

\cdark GaussianOcc~\cite{gan2024gaussianocc}
& Poses & \no
& \cdark 8.71 & \cdark 0.00 & \cdark 0.00 & \cdark 0.00 & \cdark 0.00
& \cdark 0.00 & \cdark 0.00 & \cdark 0.00 & \cdark 0.00 & \cdark 0.00 \\

\cdark GaussTR~\cite{GaussTR}
& Poses & \no
& \cdark 15.01 & \cdark 0.00 & \cdark 0.00 & \cdark 0.10 & \cdark 0.00
& \cdark 0.00 & \cdark 0.00 & \cdark 0.00 & \cdark 0.00 & \cdark 0.01 \\

\midrule
\multicolumn{12}{l}{\textit{Training-free}} \\
\midrule
\textbf{Ours (mono)} & \no & \no
& \textbf{46.81} & \textbf{19.38} & \textbf{8.88} & \textbf{38.23} & \textbf{0.21}
& \textbf{15.11}  & \textbf{12.14}  & \textbf{25.07} & \textbf{16.46}  & \textbf{16.93} \\

\textbf{Ours (rgbd)} & \no & \no
& \textbf{55.65} & \textbf{17.80} & \textbf{8.60} & \textbf{44.33} & \textbf{0.02}
& \textbf{20.76} & \textbf{16.31} & \textbf{34.90} & \textbf{24.48} & \textbf{20.90} \\
\bottomrule
\end{tabular}
\end{adjustbox}
\vspace{-15pt}
\end{table*}


\subsection{Occupancy Prediction Evaluation}
\label{sec: occ_evaluation}
We evaluate \modelname{} on EmbodiedOcc-ScanNet to assess geometric and semantic occupancy accuracy, and on the proposed ReplicaOcc to examine zero-shot generalization to unseen environments and label spaces.

\noindent\textbf{Baselines.}
We group baselines by their supervision signals and pose requirements. Results for \emph{fully supervised} methods are taken directly from the corresponding papers~\cite{embodiedocc,roboocc}. To enable a fair label-free comparison, we additionally implement two \emph{self-supervised learning methods with access to ground-truth camera poses}: GaussianOcc~\cite{gan2024gaussianocc} and GaussTR~\cite{GaussTR}. We focus on Gaussian-based baselines since Gaussian primitives provide a continuous 3D representation and can be naturally fused across frames. Both methods are originally designed for monocular inputs. We adapt them to embodied sequences by fusing per-frame Gaussian predictions into a global coordinate frame using ground-truth poses, followed by a standard Gaussian-to-occupancy conversion~\cite{gaussianformer2} to obtain voxelized occupancy maps.


\noindent\textbf{Evaluation on EmbodiedOcc-ScanNet.}
Quantitative results are reported in \cref{tab:main_embodiedocc_scannet}. As \modelname{} is the first training-free occupancy prediction framework, direct comparisons are limited. Existing self-supervised methods require \emph{ground-truth camera poses} during both training and inference and achieve IoU/mIoU scores of 10.17/4.34 and 15.63/4.95, respectively. In contrast, \modelname{} achieves 31.29/13.86 and 34.40/15.84 IoU/mIoU using monocular and RGB-D inputs, respectively, without any task-specific training. These results exceed self-supervised baselines by more than \textbf{2$\times$} across all metrics, demonstrating strong performance without annotation or learned occupancy priors.

Fully supervised methods benefit from dense 3D semantic occupancy annotations, which provide two inherent advantages during evaluation. First, EmbodiedOcc-ScanNet consolidates many object categories into coarse labels such as “objects” and “furniture”~\cite{scannet}, introducing ambiguity for open-vocabulary models not explicitly trained on this taxonomy. Second, as shown in \cref{fig: teaser}, ground-truth labels may deviate from visual evidence, leading to lower IoU/mIoU scores even when predictions better align with real-world observations.


\noindent\textbf{Zero-shot Generalization on ReplicaOcc.}
To evaluate zero-shot generalization, we directly transfer models trained on EmbodiedOcc-ScanNet to ReplicaOcc without fine-tuning or test-time adaptation. For all methods, per-frame Gaussian primitives are fused into a unified 3D coordinate frame and converted into occupancy volumes using the camera parameters and scene specifications provided by ReplicaOcc.

Quantitative results are reported in \cref{tab:replica_zeroshot_supervision}. \modelname{} exhibits strong zero-shot performance, achieving 46.81/16.93 IoU/mIoU with monocular input and further improving to 55.65/20.90 when RGB-D input is available, substantially outperforming all baselines.
In contrast, learning-based occupancy predictors fail to generalize in this transfer setting: their predictions collapse, leading to near-zero per-class scores and mIoU. This degradation can be attributed to two major domain gaps. \emph{(i) Appearance shift:} the scene geometry and visual characteristics of ScanNet differ significantly from those of Replica. \emph{(ii) Camera and scale shift:} although ground-truth poses are used at evaluation time, models trained on ScanNet tend to overfit dataset-specific camera intrinsics and metric scale, which do not transfer reliably across datasets. These results underscore the limited cross-domain generalization of both fully supervised and self-supervised learning-based occupancy predictors, which are prone to overfitting the training distribution and label space. By contrast, our training-free pipeline maintains robust geometric and semantic reasoning across environments without requiring retraining or adaptation. Qualitative comparisons are provided in Fig.~\ref{fig: occ_vis}. The above results fully validate that our approach endows occupancy prediction with generalization capabilities, which learning-based methods cannot achieve.

\begin{table}[t]
\centering
\vspace{5pt}
\caption{Geometric IoU comparison of 3DGS-based SLAM backbones for occupancy prediction on ReplicaOcc and EmbodiedOcc-ScanNet-mini.}
\label{tab:occ_iou_slam}
\footnotesize
\renewcommand{\arraystretch}{0.9}
\setlength{\tabcolsep}{5pt}
\begin{tabular}{lccc}
\toprule
\multirow{2}{*}{\textbf{Method}} 
& \multicolumn{3}{c}{\textbf{IoU}} \\
\cmidrule(lr){2-4}
& \textbf{Replica} 
& \textbf{ScanNet-mini} 
& \textbf{Average} \\

\midrule
\rowcolor{backgray}
\multicolumn{4}{l}{\textit{Monocular}} \\
\addlinespace[2pt]
Photo-SLAM~\cite{Huang_2024_CVPR}   
& 25.03      & 15.29      & 20.16 \\
MonoGS~\cite{Matsuki:Murai:etal:CVPR2024}       
& 29.50   & 14.83   & 22.17 \\
DROID-Splat~\cite{Homeyer_2025_ICCV}  
& 26.27   & 18.41   & 22.34 \\
\textbf{Ours (mono)} 
& \textbf{46.81} 
& \textbf{31.87} 
& \textbf{39.34} \\

\midrule
\rowcolor{backgray}
\multicolumn{4}{l}{\textit{RGB-D}} \\
\addlinespace[2pt]
SplaTAM~\cite{Keetha_2024_CVPR}      
& 31.11      & 17.91      & 24.51 \\
GS-ICP~\cite{10.1007/978-3-031-72764-1_11}       
& 28.95   & 21.22   & 25.09 \\
Photo-SLAM~\cite{Huang_2024_CVPR}   
& 36.97      & 16.29      & 26.63 \\
RTG-SLAM~\cite{peng2024rtgslam}      
& 35.84      & 18.46      & 27.15 \\
MonoGS~\cite{Matsuki:Murai:etal:CVPR2024}       
& 38.97   & 19.58      & 29.28 \\
DROID-Splat~\cite{Homeyer_2025_ICCV}  
& 34.48   & 24.26   & 29.37 \\
\textbf{Ours (rgbd)} 
& \textbf{55.65} 
& \textbf{34.82}
& \textbf{45.24} \\

\bottomrule
\end{tabular}
\vspace{-10pt}
\end{table}

\subsection{Ablation Study}
\label{sec: ablation_combined}
\begin{table}[t]
\centering
\vspace{5pt}
\caption{We report average results of IoU/mIoU and FPS on ReplicaOcc and EmbodiedOcc-ScanNet. \textbf{GAGU}: Geometrically Anchored Gaussian Updates. \textbf{G-ini}: Geometry-aware initialization.}
\label{tab:ablation_avg_fps}
\footnotesize
\setlength{\tabcolsep}{10pt}
\begin{tabular}{lccc}
\toprule
\textbf{Method} & \textbf{IoU} & \textbf{mIoU} & \textbf{FPS} \\
\midrule
\rowcolor{backgray}
\multicolumn{4}{l}{\textit{Monocular}} \\
w/o GAGU, G-ini & 19.88 & 10.53 & 10.7 \\
w/o G-ini       & 31.20 & 12.06 & \textbf{26.8} \\
\textbf{Ours}   & \textbf{39.05} & \textbf{15.40} & 25.3 \\
\midrule
\rowcolor{backgray}
\multicolumn{4}{l}{\textit{RGB-D}} \\
w/o GAGU, G-ini & 27.98 & 11.20 &  8.8 \\
w/o G-ini       & 40.18 & 16.03 & \textbf{25.0} \\
\textbf{Ours}   & \textbf{45.03} & \textbf{18.37} & 24.6 \\
\bottomrule
\end{tabular}
\vspace{-15pt}
\end{table}

We evaluate the effectiveness of our geometry-consistent Gaussian update strategy and the contributions of its key components (\textbf{GAGU}: geometrically anchored Gaussian updates, \textbf{G-ini}: geometry-aware initialization) on ReplicaOcc and EmbodiedOcc-ScanNet using monocular and RGB-D inputs.

\noindent\textbf{1) Comparison with 3DGS-SLAM backbones.}  
To assess geometry consistency, we compare \modelname{} against state-of-the-art 3DGS-SLAM systems~\cite{Huang_2024_CVPR,Matsuki:Murai:etal:CVPR2024,Homeyer_2025_ICCV,Keetha_2024_CVPR,10.1007/978-3-031-72764-1_11,peng2024rtgslam}. For all methods, generated 3D Gaussian maps are aligned using identical procedures (Sec.~\ref{exp_setup}) and converted into occupancy volumes (Sec.~\ref{sec: gs2occ}) to ensure fair evaluation. As reported in \cref{tab:occ_iou_slam}, \modelname{} achieves the highest geometric IoU in both RGB and RGB-D settings, with average improvements of 76.1\% and 54.0\% over the next-best method (DROID-Splat). The results demonstrate that our geometry-consistent Gaussian update strategy significantly enhances the fidelity of 3D Gaussian construction

\noindent\textbf{2) Component-wise ablation.}  
We further isolate the impact of GAGU and G-ini (Tab.~\ref{tab:ablation_avg_fps}). Removing GAGU (w/o GAGU, G-ini) significantly reduces IoU/mIoU to 19.88/10.53 (monocular) and 27.98/11.20 (RGB-D) while decreasing FPS. Introducing GAGU alone boosts IoU/mIoU to 31.20/12.06 (monocular) and 40.18/16.03 (RGB-D), improving efficiency by 1.5× and 2.8×. Adding G-ini further increases IoU/mIoU to 39.05/15.40 (monocular) and 45.03/18.37 (RGB-D) with negligible runtime loss. These results confirm that GAGU enforces long-term geometric consistency and accelerates updates, while G-ini enhances initialization for more accurate occupancy reconstruction. Together, they enable robust, real-time, and geometry-consistent 3D Gaussian mapping.

\subsection{Qualitative Results}
\label{sec: qualitative}

\begin{figure}[t]
\centering
\includegraphics[page=1, width=\linewidth]{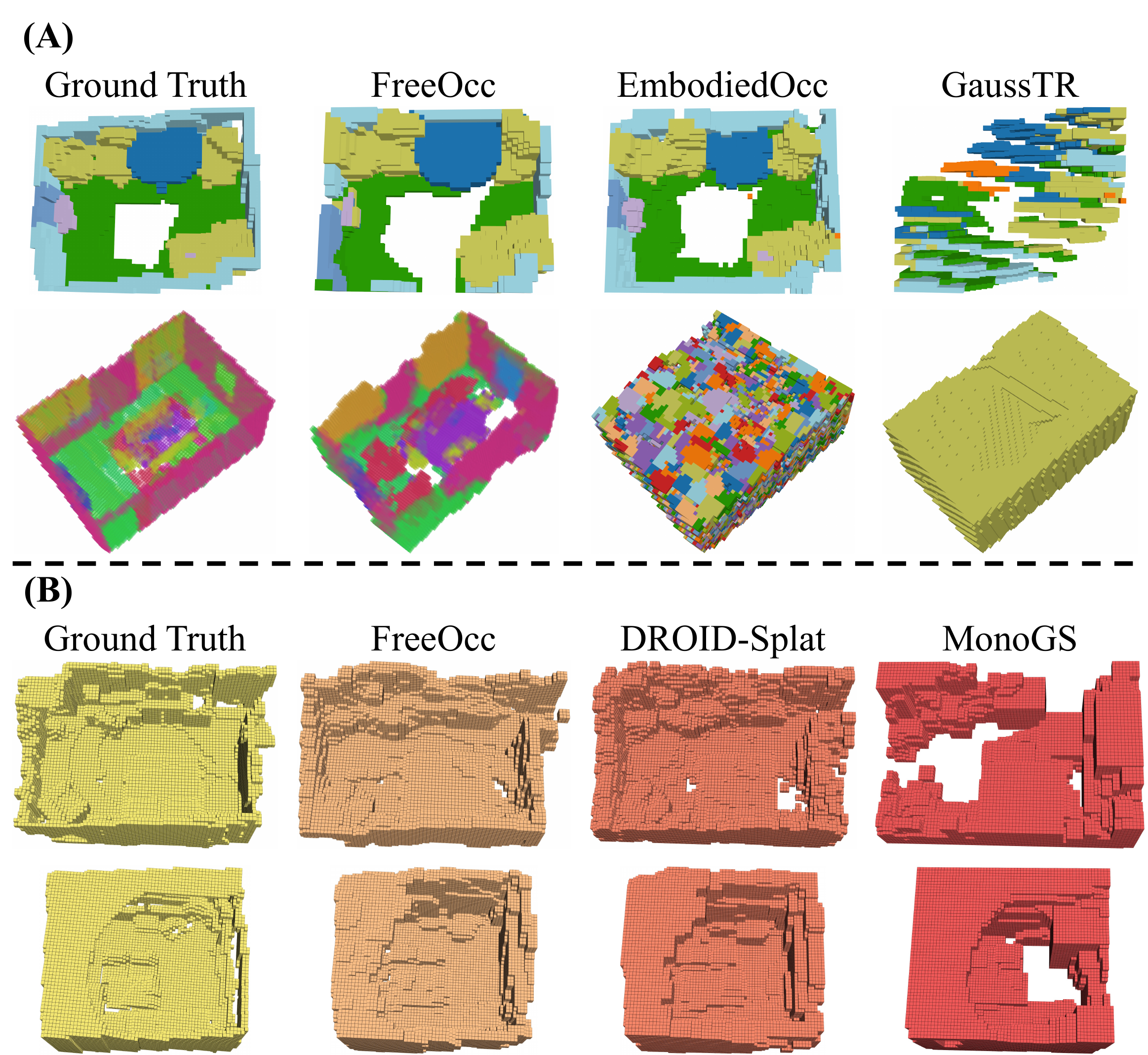} 
\caption{Qualitative occupancy prediction results. \textbf{(A)} Comparisons with learning-based occupancy predictors on “scene0470” and “room2”. \textbf{(B)} Results of the two 3DGS-SLAM methods with the highest geometric accuracy on “scene0006” and “office0”.}
\label{fig: occ_vis}
\vspace{-5pt}
\end{figure}

We present qualitative visualizations corresponding to the quantitative evaluations in \cref{sec: occ_evaluation} and \cref{sec: ablation_combined}. Representative results are shown in~\cref{fig: occ_vis}.
\cref{fig: occ_vis}(A) compares \modelname~with learning-based occupancy prediction methods on EmbodiedOcc-ScanNet and ReplicaOcc scenes. While supervised and self-supervised methods produce incomplete or near-empty occupancy maps on ReplicaOcc, \modelname{} consistently reconstructs coherent geometric structures with meaningful semantic occupancy across datasets. These results visually corroborate the strong zero-shot generalization performance observed in our quantitative analysis. Fig.~\ref{fig: occ_vis}(B) compares \modelname{} against the two 3DGS-SLAM methods with the highest geometric accuracy. Despite using similar SLAM backbones, our geometry-consistent Gaussian update strategy yields more complete and spatially consistent occupancy maps, particularly around object boundaries and thin structures, highlighting the benefits of tightly coupling Gaussian refinement with SLAM geometry.

\begin{figure}[t]
\centering
\includegraphics[page=1, width=\linewidth]{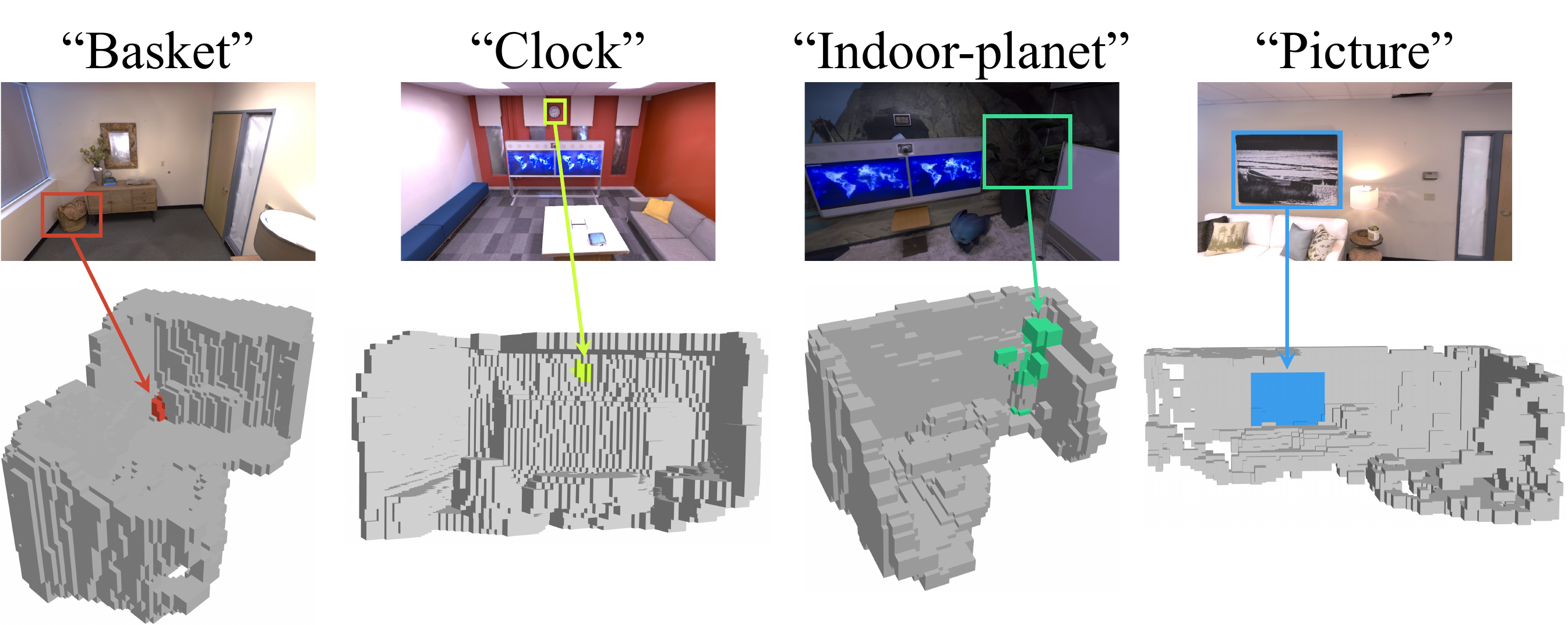} 
\caption{Open-vocabulary query results on ReplicaOcc, demonstrating semantic occupancy retrieval for different input vocabulary words.}
\label{fig: query}
\vspace{-5pt}
\end{figure}

We further visualize open-vocabulary querying results on ReplicaOcc in \cref{fig: query}. We evaluate challenging queries involving small objects (e.g., “basket” and “clock”), low-light environments (e.g., “indoor planet”), and semantically ambiguous categories (e.g., “picture”). \modelname{} successfully localizes and retrieves these targets directly from the occupancy map, demonstrating robust open-vocabulary semantic grounding in online occupancy prediction.

\section{Limitations}
\label{limitations}

Despite its strong performance, \modelname{} has several limitations. First, the quality of the resulting occupancy map is inherently tied to the robustness of the SLAM backbone. Long-term geometric and semantic consistency may degrade under accumulated drift or imperfect data association. Incorporating geometric and semantic cues derived from the occupancy representation directly into the SLAM factor graph as optimization objectives could further improve mapping robustness and consistency.
Second, current VLMs often exhibit temporal inconsistencies in semantic predictions, particularly across consecutive frames with high covisibility. Such inconsistencies introduce noise into the semantic association of Gaussian primitives. Integrating confidence-aware feature filtering or temporal consistency constraints may help mitigate these effects. We leave these directions for future work.

\section{Conclusion}
\label{sec:conclusion}

This work addresses the challenge of open-vocabulary occupancy prediction, where existing methods rely heavily on dense geometric and semantic annotations and often generalize poorly to novel environments. We introduced \modelname{}, the first training-free framework for open-vocabulary occupancy prediction in embodied indoor settings. By designing a multi-layer mapping pipeline that tightly integrates SLAM geometry, 3D Gaussian representations, and vision–language semantics, \modelname{} constructs globally consistent occupancy maps without task-specific training or voxel-level supervision.
Extensive experiments demonstrate that \modelname{} achieves competitive geometric and semantic accuracy while significantly improving generalization compared to learning-based approaches. We believe this work represents an important step toward scalable, annotation-free occupancy prediction, enabling robots to reason about arbitrary 3D environments and semantics in real-world deployments.

\section*{Acknowledgments}
This work was supported by National Natural Science Foundation of China (No.62573370), the Key Area Project of Education Department of Guangdong Province (No.2025ZDZX3051) and Guangdong Provincial Key Lab of Integrated Communication, Sensing and Computation for Ubiquitous Internet of Things (No.2023B1212010007).

\bibliographystyle{plainnat}
\bibliography{references}

@article{pop3d,
  title={Pop-3d: Open-vocabulary 3d occupancy prediction from images},
  author={Vobecky, Antonin and Sim{\'e}oni, Oriane and Hurych, David and Gidaris, Spyridon and Bursuc, Andrei and P{\'e}rez, Patrick and Sivic, Josef},
  journal={Advances in Neural Information Processing Systems},
  volume={36},
  pages={50545--50557},
  year={2023}
}

@InProceedings{embodiedocc,
    author    = {Wu, Yuqi and Zheng, Wenzhao and Zuo, Sicheng and Huang, Yuanhui and Zhou, Jie and Lu, Jiwen},
    title     = {EmbodiedOcc: Embodied 3D Occupancy Prediction for Vision-based Online Scene Understanding},
    booktitle = {Proceedings of the IEEE/CVF International Conference on Computer Vision (ICCV)},
    month     = {October},
    year      = {2025},
    pages     = {26360-26370}
}

@inproceedings{embodiedocc++,
author = {Wang, Hao and Wei, Xiaobao and Zhang, Xiaoan and Li, Jianing and Bai, Chengyu and Li, Ying and Lu, Ming and Zheng, Wenzhao and Zhang, Shanghang},
title = {EmbodiedOcc++: Boosting Embodied 3D Occupancy Prediction with Plane Regularization and Uncertainty Sampler},
year = {2025},
publisher = {Association for Computing Machinery},
address = {New York, NY, USA},
booktitle = {Proceedings of the 33rd ACM International Conference on Multimedia},
pages = {925–934},
numpages = {10},
location = {Dublin, Ireland},
series = {MM '25}
}

@inproceedings{gaussianformer,
  title={Gaussianformer: Scene as gaussians for vision-based 3d semantic occupancy prediction},
  author={Huang, Yuanhui and Zheng, Wenzhao and Zhang, Yunpeng and Zhou, Jie and Lu, Jiwen},
  booktitle={European Conference on Computer Vision},
  pages={376--393},
  year={2024},
  organization={Springer}
}

@InProceedings{gaussianformer2,
    author    = {Huang, Yuanhui and Thammatadatrakoon, Amonnut and Zheng, Wenzhao and Zhang, Yunpeng and Du, Dalong and Lu, Jiwen},
    title     = {GaussianFormer-2: Probabilistic Gaussian Superposition for Efficient 3D Occupancy Prediction},
    booktitle = {Proceedings of the IEEE/CVF Conference on Computer Vision and Pattern Recognition (CVPR)},
    year      = {2025},
    pages     = {27477-27486}
}

@inproceedings{monoscene,
  title={Monoscene: Monocular 3d semantic scene completion},
  author={Cao, Anh-Quan and De Charette, Raoul},
  booktitle={Proceedings of the IEEE/CVF Conference on Computer Vision and Pattern Recognition},
  pages={3991--4001},
  year={2022}
}

@inproceedings{ISO,
  title={Monocular occupancy prediction for scalable indoor scenes},
  author={Yu, Hongxiao and Wang, Yuqi and Chen, Yuntao and Zhang, Zhaoxiang},
  booktitle={European Conference on Computer Vision},
  pages={38--54},
  year={2024},
  organization={Springer}
}

@inproceedings{surroundocc,
  title={Surroundocc: Multi-camera 3d occupancy prediction for autonomous driving},
  author={Wei, Yi and Zhao, Linqing and Zheng, Wenzhao and Zhu, Zheng and Zhou, Jie and Lu, Jiwen},
  booktitle={Proceedings of the IEEE/CVF International Conference on Computer Vision},
  pages={21729--21740},
  year={2023}
}

@inproceedings{voxformer,
  title={Voxformer: Sparse voxel transformer for camera-based 3d semantic scene completion},
  author={Li, Yiming and Yu, Zhiding and Choy, Christopher and Xiao, Chaowei and Alvarez, Jose M and Fidler, Sanja and Feng, Chen and Anandkumar, Anima},
  booktitle={Proceedings of the IEEE/CVF conference on computer vision and pattern recognition},
  pages={9087--9098},
  year={2023}
}

@inproceedings{occupancypoints,
  title={Occupancy as set of points},
  author={Shi, Yiang and Cheng, Tianheng and Zhang, Qian and Liu, Wenyu and Wang, Xinggang},
  booktitle={European Conference on Computer Vision},
  pages={72--87},
  year={2024},
  organization={Springer}
}

@inproceedings{lss,
  title={Lift, splat, shoot: Encoding images from arbitrary camera rigs by implicitly unprojecting to 3d},
  author={Philion, Jonah and Fidler, Sanja},
  booktitle={European conference on computer vision},
  pages={194--210},
  year={2020},
  organization={Springer}
}

@inproceedings{Triformer,
  title={Tri-perspective view for vision-based 3d semantic occupancy prediction},
  author={Huang, Yuanhui and Zheng, Wenzhao and Zhang, Yunpeng and Zhou, Jie and Lu, Jiwen},
  booktitle={Proceedings of the IEEE/CVF conference on computer vision and pattern recognition},
  pages={9223--9232},
  year={2023}
}

@inproceedings{scannet,
  title={Scannet: Richly-annotated 3d reconstructions of indoor scenes},
  author={Dai, Angela and Chang, Angel X and Savva, Manolis and Halber, Maciej and Funkhouser, Thomas and Nie{\ss}ner, Matthias},
  booktitle={Proceedings of the IEEE conference on computer vision and pattern recognition},
  pages={5828--5839},
  year={2017}
}

@inproceedings{ndcscene,
  title={Ndc-scene: Boost monocular 3d semantic scene completion in normalized device coordinates space},
  author={Yao, Jiawei and Li, Chuming and Sun, Keqiang and Cai, Yingjie and Li, Hao and Ouyang, Wanli and Li, Hongsheng},
  booktitle={2023 IEEE/CVF International Conference on Computer Vision},
  pages={9421--9431},
  year={2023}
}

@inproceedings{sscnet,
  title={Semantic scene completion from a single depth image},
  author={Song, Shuran and Yu, Fisher and Zeng, Andy and Chang, Angel X and Savva, Manolis and Funkhouser, Thomas},
  booktitle={Proceedings of the IEEE conference on computer vision and pattern recognition},
  pages={1746--1754},
  year={2017}
}

@inproceedings{sparseocc,
  title={Sparseocc: Rethinking sparse latent representation for vision-based semantic occupancy prediction},
  author={Tang, Pin and Wang, Zhongdao and Wang, Guoqing and Zheng, Jilai and Ren, Xiangxuan and Feng, Bailan and Ma, Chao},
  booktitle={Proceedings of the IEEE/CVF Conference on Computer Vision and Pattern Recognition},
  pages={15035--15044},
  year={2024}
}

@article{fbocc,
  title={{FB-OCC}: {3D} Occupancy Prediction based on Forward-Backward View Transformation},
  author={Li, Zhiqi and Yu, Zhiding and Austin, David and Fang, Mingsheng and Lan, Shiyi and Kautz, Jan and Alvarez, Jose M},
  journal={arXiv:2307.01492},
  year={2023}
}

@article{octreeocc,
  title={Octreeocc: Efficient and multi-granularity occupancy prediction using octree queries},
  author={Lu, Yuhang and Zhu, Xinge and Wang, Tai and Ma, Yuexin},
  journal={Advances in Neural Information Processing Systems},
  volume={37},
  pages={79618--79641},
  year={2024}
}

@article{gaussiansplatting,
  title={3D Gaussian splatting for real-time radiance field rendering.},
  author={Kerbl, Bernhard and Kopanas, Georgios and Leimk{\"u}hler, Thomas and Drettakis, George},
  journal={ACM Trans. Graph.},
  volume={42},
  number={4},
  pages={139--1},
  year={2023}
}

@article{gao2025loc,
  title={LOC: A General Language-Guided Framework for Open-Set 3D Occupancy Prediction},
  author={Gao, Yuhang and Xiang, Xiang and Zhong, Sheng and Wang, Guoyou},
  journal={arXiv preprint arXiv:2510.22141},
  year={2025}
}

@InProceedings{li2025ago,
    author    = {Li, Peizheng and Ding, Shuxiao and Zhou, You and Zhang, Qingwen and Inak, Onat and Triess, Larissa and Hanselmann, Niklas and Cordts, Marius and Zell, Andreas},
    title     = {AGO: Adaptive Grounding for Open World 3D Occupancy Prediction},
    booktitle = {Proceedings of the IEEE/CVF International Conference on Computer Vision},
    month     = {October},
    year      = {2025},
    pages     = {8645-8655}
}

@article{occnerf,
  publtype={informal},
  author={Chubin Zhang and Juncheng Yan and Yi Wei and Jiaxin Li and Li Liu and Yansong Tang and Yueqi Duan and Jiwen Lu},
  title={OccNeRF: Self-Supervised Multi-Camera Occupancy Prediction with Neural Radiance Fields},
  year={2023},
  cdate={1672531200000},
  journal={CoRR},
  volume={abs/2312.09243},
}

@inproceedings{
boeder2024langocc,
title={LangOcc: Open Vocabulary Occupancy Estimation via Volume Rendering},
author={Simon Boeder and Fabian Gigengack and Benjamin Risse},
booktitle={International Conference on 3D Vision 2025},
year={2025},
}

@inproceedings{veon,
  title={Veon: Vocabulary-enhanced occupancy prediction},
  author={Zheng, Jilai and Tang, Pin and Wang, Zhongdao and Wang, Guoqing and Ren, Xiangxuan and Feng, Bailan and Ma, Chao},
  booktitle={European Conference on Computer Vision},
  pages={92--108},
  year={2024},
  organization={Springer}
}

@inproceedings{clip,
  title={Learning transferable visual models from natural language supervision},
  author={Radford, Alec and Kim, Jong Wook and Hallacy, Chris and Ramesh, Aditya and Goh, Gabriel and Agarwal, Sandhini and Sastry, Girish and Askell, Amanda and Mishkin, Pamela and Clark, Jack and others},
  booktitle={International conference on machine learning},
  pages={8748--8763},
  year={2021},
  organization={PmLR}
}

@InProceedings{selfocc,
    author    = {Huang, Yuanhui and Zheng, Wenzhao and Zhang, Borui and Zhou, Jie and Lu, Jiwen},
    title     = {SelfOcc: Self-Supervised Vision-Based 3D Occupancy Prediction},
    booktitle = {Proceedings of the IEEE/CVF Conference on Computer Vision and Pattern Recognition},
    year      = {2024}
}

@inproceedings{lv2021clins,
  title={CLINS: Continuous-Time Trajectory Estimation for LiDAR-Inertial System},
  author={Lv, Jiajun and Hu, Kewei and Xu, Jinhong and Liu, Yong and Ma, Xiushui and Zuo, Xingxing},
  booktitle={2021 IEEE/RSJ International Conference on Intelligent Robots and Systems},
  pages={6657--6663},
  year={2021},
  organization={IEEE}
}

@inproceedings{lang2025gaussian,
  title={Gaussian-LIC: Real-Time Photo-Realistic SLAM with Gaussian Splatting and LiDAR-Inertial-Camera Fusion}, 
  author={Lang, Xiaolei and Li, Laijian and Wu, Chenming and Zhao, Chen and Liu, Lina and Liu, Yong and Lv, Jiajun and Zuo, Xingxing},
  booktitle={2025 International Conference on Robotics and Automation},
  year={2025},
  organization={IEEE}
}

@article{lang2025gaussianlic2,
  title={Gaussian-lic2: Lidar-inertial-camera gaussian splatting slam},
  author={Lang, Xiaolei and Lv, Jiajun and Tang, Kai and Li, Laijian and Huang, Jianxin and Liu, Lina and Liu, Yong and Zuo, Xingxing},
  journal={arXiv preprint arXiv:2507.04004},
  year={2025}
}

@article{li2025pg,
  title={PG-SLAM: Photo-realistic and geometry-aware RGB-D SLAM in dynamic environments},
  author={Li, Haoang and Meng, Xiangqi and Zuo, Xingxing and Liu, Zhe and Wang, Hesheng and Cremers, Daniel},
  journal={IEEE Transactions on Robotics},
  year={2025},
  publisher={IEEE}
}

@article{trident,
    title={Harnessing Vision Foundation Models for High-Performance, Training-Free Open Vocabulary Segmentation},
    author={Yuheng Shi and Minjing Dong and Chang Xu},
    journal={arXiv preprint arXiv:2411.09219},
    year={2024},
}

@article{bai2025qwen3,
  title={Qwen3-vl technical report},
  author={Bai, Shuai and Cai, Yuxuan and Chen, Ruizhe and Chen, Keqin and Chen, Xionghui and Cheng, Zesen and Deng, Lianghao and Ding, Wei and Gao, Chang and Ge, Chunjiang and others},
  journal={arXiv preprint arXiv:2511.21631},
  year={2025}
}

@article{roboocc,
  title={Roboocc: Enhancing the geometric and semantic scene understanding for robots},
  author={Zhang, Zhang and Zhang, Qiang and Cui, Wei and Shi, Shuai and Guo, Yijie and Han, Gang and Zhao, Wen and Ren, Hengle and Xu, Renjing and Tang, Jian},
  journal={arXiv preprint arXiv:2504.14604},
  year={2025}
}

@inproceedings{pan2024renderocc,
  title={Renderocc: Vision-centric 3d occupancy prediction with 2d rendering supervision},
  author={Pan, Mingjie and Liu, Jiaming and Zhang, Renrui and Huang, Peixiang and Li, Xiaoqi and Xie, Hongwei and Wang, Bing and Liu, Li and Zhang, Shanghang},
  booktitle={2024 IEEE International Conference on Robotics and Automation},
  pages={12404--12411},
  year={2024},
  organization={IEEE}
}

@inproceedings{liu2024gausstrfoundationmodelaligned,
  title={Gausstr: Foundation model-aligned gaussian transformer for self-supervised 3d spatial understanding},
  author={Jiang, Haoyi and Liu, Liu and Cheng, Tianheng and Wang, Xinjie and Lin, Tianwei and Su, Zhizhong and Liu, Wenyu and Wang, Xinggang},
  booktitle={Proceedings of the Computer Vision and Pattern Recognition Conference},
  year={2025}
}

@misc{boeder2025gaussianflowoccsparseweakly,
      title={GaussianFlowOcc: Sparse and Weakly Supervised Occupancy Estimation using  Gaussian Splatting and Temporal Flow}, 
      author={Simon Boeder and Fabian Gigengack and Benjamin Risse},
      year={2025},
      eprint={2502.17288},
      archivePrefix={arXiv},
      primaryClass={cs.CV},
}

@InProceedings{Yan_2024_CVPR,
    author    = {Yan, Chi and Qu, Delin and Xu, Dan and Zhao, Bin and Wang, Zhigang and Wang, Dong and Li, Xuelong},
    title     = {GS-SLAM: Dense Visual SLAM with 3D Gaussian Splatting},
    booktitle = {Proceedings of the IEEE/CVF Conference on Computer Vision and Pattern Recognition},
    year      = {2024}
}

@InProceedings{Huang_2024_CVPR,
    author    = {Huang, Huajian and Li, Longwei and Cheng, Hui and Yeung, Sai-Kit},
    title     = {Photo-SLAM: Real-time Simultaneous Localization and Photorealistic Mapping for Monocular Stereo and RGB-D Cameras},
    booktitle = {Proceedings of the IEEE/CVF Conference on Computer Vision and Pattern Recognition},
    year      = {2024}
}

@InProceedings{Keetha_2024_CVPR,
    author    = {Keetha, Nikhil and Karhade, Jay and Jatavallabhula, Krishna Murthy and Yang, Gengshan and Scherer, Sebastian and Ramanan, Deva and Luiten, Jonathon},
    title     = {SplaTAM: Splat Track \& Map 3D Gaussians for Dense RGB-D SLAM},
    booktitle = {Proceedings of the IEEE/CVF Conference on Computer Vision and Pattern Recognition},
    year      = {2024}
}

@inproceedings{Matsuki:Murai:etal:CVPR2024,
  title={{G}aussian {S}platting {SLAM}},
  author={Hidenobu Matsuki and Riku Murai and Paul H. J. Kelly and Andrew J. Davison},
  booktitle={Proceedings of the IEEE/CVF Conference on Computer Vision and Pattern Recognition},
  year={2024}
}

@misc{yugay2023gaussianslam,
      title={Gaussian-SLAM: Photo-realistic Dense SLAM with Gaussian Splatting}, 
      author={Vladimir Yugay and Yue Li and Theo Gevers and Martin R. Oswald},
      year={2023},
      eprint={2312.10070},
      archivePrefix={arXiv},
      primaryClass={cs.CV}
}

@Article{kerbl3Dgaussians,
      author       = {Kerbl, Bernhard and Kopanas, Georgios and Leimk{\"u}hler, Thomas and Drettakis, George},
      title        = {3D Gaussian Splatting for Real-Time Radiance Field Rendering},
      journal      = {ACM Transactions on Graphics},
      number       = {4},
      volume       = {42},
      month        = {July},
      year         = {2023}
}

@inproceedings{li2024sgs,
  title={Sgs-slam: Semantic gaussian splatting for neural dense slam},
  author={Li, Mingrui and Liu, Shuhong and Zhou, Heng and Zhu, Guohao and Cheng, Na and Deng, Tianchen and Wang, Hongyu},
  booktitle={European Conference on Computer Vision},
  pages={163--179},
  year={2024},
  organization={Springer}
}

@article{Ji_2024,
   title={NEDS-SLAM: A Neural Explicit Dense Semantic SLAM Framework Using 3D Gaussian Splatting},
   volume={9},
   ISSN={2377-3774},
   DOI={10.1109/lra.2024.3451390},
   number={10},
   journal={IEEE Robotics and Automation Letters},
   publisher={Institute of Electrical and Electronics Engineers (IEEE)},
   author={Ji, Yiming and Liu, Yang and Xie, Guanghu and Ma, Boyu and Xie, Zongwu and Liu, Hong},
   year={2024},
   month=oct, pages={8778–8785} }

@inproceedings{li2024gs3lam,
      author = {Li, Linfei and Zhang, Lin and Wang, Zhong and Shen, Ying},
      title = {GS3LAM: Gaussian Semantic Splatting SLAM},
      year = {2024},
      publisher = {Association for Computing Machinery},
      address = {New York, NY, USA},
      booktitle = {Proceedings of the 32nd ACM International Conference on Multimedia},
      pages = {3019–3027},
      numpages = {9},
      location = {Melbourne VIC, Australia},
      series = {MM '24}
}

@misc{yang2025opengsslamopensetdensesemantic,
      title={OpenGS-SLAM: Open-Set Dense Semantic SLAM with 3D Gaussian Splatting for Object-Level Scene Understanding}, 
      author={Dianyi Yang and Yu Gao and Xihan Wang and Yufeng Yue and Yi Yang and Mengyin Fu},
      year={2025},
      eprint={2503.01646},
      archivePrefix={arXiv},
      primaryClass={cs.CV},
}

@inproceedings{wang2023semantic,
  title={Semantic scene completion with cleaner self},
  author={Wang, Fengyun and Zhang, Dong and Zhang, Hanwang and Tang, Jinhui and Sun, Qianru},
  booktitle={Proceedings of the IEEE/CVF Conference on Computer Vision and Pattern Recognition},
  pages={867--877},
  year={2023}
}

@InProceedings{gan2024gaussianocc,
    author    = {Gan, Wanshui and Liu, Fang and Xu, Hongbin and Mo, Ningkai and Yokoya, Naoto},
    title     = {GaussianOcc: Fully Self-supervised and Efficient 3D Occupancy Estimation with Gaussian Splatting},
    booktitle = {Proceedings of the IEEE/CVF International Conference on Computer Vision},
    year      = {2025}
}

@inproceedings{GaussTR,
    title     = {GaussTR: Foundation Model-Aligned Gaussian Transformer for Self-Supervised 3D Spatial Understanding},
    author    = {Haoyi Jiang and Liu Liu and Tianheng Cheng and Xinjie Wang and Tianwei Lin and Zhizhong Su and Wenyu Liu and Xinggang Wang},
    year      = {2025},
    booktitle = {Proceedings of the IEEE/CVF Conference on Computer Vision and Pattern Recognition}
}

@article{teed2021droid,
  title={{DROID-SLAM: Deep Visual SLAM for Monocular, Stereo, and RGB-D Cameras}},
  author={Teed, Zachary and Deng, Jia},
  journal={Advances in neural information processing systems},
  year={2021}
}

@inproceedings{murai2024_mast3rslam,
  title={{MASt3R-SLAM}: Real-Time Dense {SLAM} with {3D} Reconstruction Priors},
  author={Murai, Riku and Dexheimer, Eric and Davison, Andrew J.},
  booktitle={Proceedings of the IEEE/CVF Conference on Computer Vision and Pattern Recognition},
  year={2025},
}

@article{maggio2025vggt-slam,
  title={VGGT-SLAM: Dense RGB SLAM Optimized on the SL (4) Manifold},
  author={Maggio, Dominic and Lim, Hyungtae and Carlone, Luca},
  journal={Advances in Neural Information Processing Systems},
  volume={39},
  year={2025}
}

@inproceedings{schoenberger2016sfm,
    author={Sch\"{o}nberger, Johannes Lutz and Frahm, Jan-Michael},
    title={Structure-from-Motion Revisited},
    booktitle={Proceedings of the IEEE/CVF Conference on Computer Vision and Pattern Recognition},
    year={2016}
}

@inproceedings{624b677c512440f9816f866607d16018,
  title={RAFT: Recurrent All-Pairs Field Transforms for Optical Flow},
  author={Zachary Teed and Jia Deng},
  booktitle={European Conference on Computer Vision},
  pages={402--419},
  year={2020},
  organization={Springer}
}

@InProceedings{Homeyer_2025_ICCV,
    author    = {Homeyer, Christian and Begiristain, Leon and Schn\"orr, Christoph},
    title     = {DROID-Splat Combining end-to-end SLAM with 3D Gaussian Splatting},
    booktitle = {Proceedings of the IEEE/CVF International Conference on Computer Vision Workshops},
    year      = {2025}
}

@inproceedings{brachmann2023ace,
    title={Accelerated Coordinate Encoding: Learning to Relocalize in Minutes using RGB and Poses},
    author={Brachmann, Eric and Cavallari, Tommaso and Prisacariu, Victor Adrian},
    booktitle={Proceedings of the IEEE/CVF Conference on Computer Vision and Pattern Recognition},
    year={2023},
}

@article{sun2021loftr,
  title={{LoFTR}: Detector-Free Local Feature Matching with Transformers},
  author={Sun, Jiaming and Shen, Zehong and Wang, Yuang and Bao, Hujun and Zhou, Xiaowei},
  journal={Proceedings of the IEEE/CVF Conference on Computer Vision and Pattern Recognition},
  year={2021}
}

@article{straub2019replica,
  title={The Replica dataset: A digital replica of indoor spaces},
  author={Straub, Julian and Whelan, Thomas and Ma, Lingni and Chen, Yufan and Wijmans, Erik and Green, Simon and Engel, Jakob J and Mur-Artal, Raul and Ren, Carl and Verma, Shobhit and others},
  journal={arXiv preprint arXiv:1906.05797},
  year={2019}
}

@inproceedings{Zhu2022CVPR,
  author    = {Zhu, Zihan and Peng, Songyou and Larsson, Viktor and Xu, Weiwei and Bao, Hujun and Cui, Zhaopeng and Oswald, Martin R. and Pollefeys, Marc},
  title     = {NICE-SLAM: Neural Implicit Scalable Encoding for SLAM},
  booktitle = {Proceedings of the IEEE/CVF Conference on Computer Vision and Pattern Recognition},
  year      = {2022}
}

@inproceedings{dai2017scannet,
    title={ScanNet: Richly-annotated 3D Reconstructions of Indoor Scenes},
    author={Dai, Angela and Chang, Angel X. and Savva, Manolis and Halber, Maciej and Funkhouser, Thomas and Nie{\ss}ner, Matthias},
    booktitle = {Proceedings of the IEEE/CVF Conference on Computer Vision and Pattern Recognition},
    year = {2017}
}

@misc{grupp2017evo,
  title={evo: Python package for the evaluation of odometry and SLAM.},
  author={Grupp, Michael},
  howpublished={\url{https://github.com/MichaelGrupp/evo}},
  year={2017}
}

@inproceedings{10.1007/978-3-031-72764-1_11,
  title={RGBD GS-ICP SLAM},
  author={Ha, Seongbo and Yeon, Jiung and Yu, Hyeonwoo},
  booktitle={European Conference on Computer Vision},
  pages={180–197},
  year={2020},
  organization={Springer}
}

@article{peng2024rtgslam,
        title = {RTG-SLAM: Real-time 3D Reconstruction at Scale using Gaussian Splatting},
        author = {Zhexi Peng and Tianjia Shao and Liu Yong and Jingke Zhou and Yin Yang and Jingdong Wang and Kun Zhou},
        journal = {ACM SIGGRAPH Conference Proceedings, Denver, CO, United States, July 28 - August 1, 2024},
        year = {2024}
      }

@article{murORB2,
  title={{ORB-SLAM2}: an Open-Source {SLAM} System for Monocular, Stereo and {RGB-D} Cameras},
  author={Mur-Artal, Ra\'ul and Tard\'os, Juan D.},
  journal={IEEE Transactions on Robotics},
  volume={33},
  number={5},
  pages={1255--1262},
  doi = {10.1109/TRO.2017.2705103},
  year={2017}
 }

@INPROCEEDINGS{11127380,
  author={Hu, Yan Song and Abboud, Nicolas and Ali, Muhammad Qasim and Yang, Adam Srebrnjak and Elhajj, Imad and Asmar, Daniel and Chen, Yuhao and Zelek, John S.},
  booktitle={2025 IEEE International Conference on Robotics and Automation (ICRA)}, 
  title={MGSO: Monocular Real-Time Photometric SLAM with Efficient 3D Gaussian Splatting}, 
  year={2025},
  volume={},
  number={},
  pages={11061-11067},
  doi={10.1109/ICRA55743.2025.11127380}}

@article{10.1145/3503250,
author = {Mildenhall, Ben and Srinivasan, Pratul P. and Tancik, Matthew and Barron, Jonathan T. and Ramamoorthi, Ravi and Ng, Ren},
title = {NeRF: representing scenes as neural radiance fields for view synthesis},
year = {2021},
issue_date = {January 2022},
publisher = {Association for Computing Machinery},
address = {New York, NY, USA},
volume = {65},
number = {1},
issn = {0001-0782},
doi = {10.1145/3503250},
journal = {Commun. ACM},
month = dec,
pages = {99–106},
numpages = {8}
}

@misc{deng2026best3dscenerepresentation,
      title={What Is The Best 3D Scene Representation for Robotics? From Geometric to Foundation Models}, 
      author={Tianchen Deng and Yue Pan and Shenghai Yuan and Dong Li and Chen Wang and Mingrui Li and Long Chen and Lihua Xie and Danwei Wang and Jingchuan Wang and Javier Civera and Hesheng Wang and Weidong Chen},
      year={2026},
      eprint={2512.03422},
      archivePrefix={arXiv},
      primaryClass={cs.RO},
}

@article{ullman1979interpretation,
  title={The interpretation of structure from motion},
  author={Ullman, Shimon},
  journal={Proceedings of the Royal Society of London. Series B. Biological Sciences},
  volume={203},
  number={1153},
  pages={405--426},
  year={1979},
  publisher={The Royal Society London}
}

@article{chen1989representation,
  title={Representation, display, and manipulation of 3D digital scenes and their medical applications},
  author={Chen, Lih-Shyang and Sontag, Marc R},
  journal={Computer Vision, Graphics, and Image Processing},
  volume={48},
  number={2},
  pages={190--216},
  year={1989},
  publisher={Elsevier}
}

@article{cadena2017past,
  title={Past, present, and future of simultaneous localization and mapping: Toward the robust-perception age},
  author={Cadena, Cesar and Carlone, Luca and Carrillo, Henry and Latif, Yasir and Scaramuzza, Davide and Neira, Jos{\'e} and Reid, Ian and Leonard, John J},
  journal={IEEE Transactions on robotics},
  volume={32},
  number={6},
  pages={1309--1332},
  year={2017},
  publisher={IEEE}
}

@article{fei20243d,
  title={3d gaussian splatting as new era: A survey},
  author={Fei, Ben and Xu, Jingyi and Zhang, Rui and Zhou, Qingyuan and Yang, Weidong and He, Ying},
  journal={IEEE Transactions on Visualization and Computer Graphics},
  year={2024},
  publisher={IEEE}
}

@article{li2025enhancing,
  title={Enhancing Indoor Occupancy Prediction via Sparse Query-Based Multi-Level Consistent Knowledge Distillation},
  author={Li, Xiang and Zheng, Yupeng and Li, Pengfei and Chen, Yilun and Zhang, Ya-Qin and Ding, Wenchao},
  journal={IEEE Robotics and Automation Letters},
  year={2025},
  publisher={IEEE}
}

@article{frieder1985back,
  title={Back-to-front display of voxel based objects},
  author={Frieder, Gideon and Gordon, Dan and Reynolds, R},
  journal={IEEE Computer Graphics and Applications},
  volume={5},
  number={01},
  pages={52--60},
  year={1985},
  publisher={IEEE Computer Society}
}

@misc{poggi2024selfevolvingdepthsupervised3d,
      title={Self-Evolving Depth-Supervised 3D Gaussian Splatting from Rendered Stereo Pairs}, 
      author={Matteo Poggi and Fabio Tosi and Fatma Güney and Sadra Safadoust},
      year={2024},
      eprint={2409.07456},
      archivePrefix={arXiv},
      primaryClass={cs.CV},
}

@inproceedings{ververas2024sags,
  title={SAGS: Structure-Aware 3D Gaussian splatting},
  author={Ververas, Evangelos and Potamias, Rolandos Alexandros and Song, Jifei and Deng, Jiankang and Zafeiriou, Stefanos},
  booktitle={European Conference on Computer Vision},
  pages={221--238},
  year={2024},
  organization={Springer}
}

@ARTICLE{hornung13auro,
  author = {Armin Hornung and Kai M. Wurm and Maren Bennewitz and Cyrill
  Stachniss and Wolfram Burgard},
  title = {{OctoMap}: An Efficient Probabilistic {3D} Mapping Framework Based
  on Octrees},
  journal = {Autonomous Robots},
  year = 2013,
  doi = {10.1007/s10514-012-9321-0}
}

@article{song2016ssc,
  author     = {Song, Shuran and Yu, Fisher  and Zeng, Andy and Chang, Angel X and Savva, Manolis and Funkhouser, Thomas},
  title      = {Semantic Scene Completion from a Single Depth Image},
  journal    = {arXiv preprint arXiv:1611.08974},
  year       = {2016},
}

@inproceedings{zhou2024feature,
  title={Feature 3dgs: Supercharging 3d gaussian splatting to enable distilled feature fields},
  author={Zhou, Shijie and Chang, Haoran and Jiang, Sicheng and Fan, Zhiwen and Zhu, Zehao and Xu, Dejia and Chari, Pradyumna and You, Suya and Wang, Zhangyang and Kadambi, Achuta},
  booktitle={Proceedings of the IEEE/CVF Conference on Computer Vision and Pattern Recognition},
  pages={21676--21685},
  year={2024}
}

@inproceedings{shi2024language,
  title={Language embedded 3d gaussians for open-vocabulary scene understanding},
  author={Shi, Jin-Chuan and Wang, Miao and Duan, Hao-Bin and Guan, Shao-Hua},
  booktitle={Proceedings of the IEEE/CVF Conference on Computer Vision and Pattern Recognition},
  pages={5333--5343},
  year={2024}
}

@article{lange2024self,
  author     = {Bernard Lange and Masha Itkina and Jiachen Li and Mykel Kochenderfer},
  title      = {Self-supervised Multi-future Occupancy Forecasting for Autonomous Driving},
  journal    = {Robotics: Science and Systems},
  year       = {2025},
}

@article{pan2025pings,
  author     = {Yue Pan and Xingguang Zhong and Liren Jin and Louis Wiesmann and Marija Popovic and Jens Behley and Cyrill Stachniss},
  title      = {PINGS: Gaussian Splatting Meets Distance Fields within a Point-Based Implicit Neural Map},
  journal    = {Robotics: Science and Systems},
  year       = {2025},
}

@article{roman,
  author     = {Mason Peterson and Yixuan Jia and Yulun Tian and Annika Thomas and Jonathan P. How},
  title      = {Roman: Open-set object map alignment for robust view-invariant global localization},
  journal    = {Robotics: Science and Systems},
  year       = {2025},
}

@article{zhou2026generalizing,
  title={Generalizing Visual Geometry Priors to Sparse Gaussian Occupancy Prediction},
  author={Zhou, Changqing and Luo, Yueru and Chen, Changhao},
  journal={arXiv preprint arXiv:2602.21552},
  year={2026}
}

@article{zhou2026monocular,
  title={Monocular Open Vocabulary Occupancy Prediction for Indoor Scenes},
  author={Zhou, Changqing and Luo, Yueru and Zhang, Han and Jiang, Zeyu and Chen, Changhao},
  journal={arXiv preprint arXiv:2602.22667},
  year={2026}
}

@inproceedings{
zou2023segment,
title={Segment Everything Everywhere All at Once},
author={Xueyan Zou and Jianwei Yang and Hao Zhang and Feng Li and Linjie Li and Jianfeng Wang and Lijuan Wang and Jianfeng Gao and Yong Jae Lee},
booktitle={Thirty-seventh Conference on Neural Information Processing Systems},
year={2023},
}

@article{
oquab2024dinov,
title={{DINO}v2: Learning Robust Visual Features without Supervision},
author={Maxime Oquab and Timoth{\'e}e Darcet and Th{\'e}o Moutakanni and Huy V. Vo and Marc Szafraniec and Vasil Khalidov and Pierre Fernandez and Daniel HAZIZA and Francisco Massa and Alaaeldin El-Nouby and Mido Assran and Nicolas Ballas and Wojciech Galuba and Russell Howes and Po-Yao Huang and Shang-Wen Li and Ishan Misra and Michael Rabbat and Vasu Sharma and Gabriel Synnaeve and Hu Xu and Herve Jegou and Julien Mairal and Patrick Labatut and Armand Joulin and Piotr Bojanowski},
journal={Transactions on Machine Learning Research},
issn={2835-8856},
year={2024},
url={https://openreview.net/forum?id=a68SUt6zFt},
note={Featured Certification}
}

@inproceedings{fischedick2025efficient,
  title={Efficient Prediction of Dense Visual Embeddings via Distillation and RGB-D Transformers},
  author={Fischedick, S{\"o}hnke Benedikt and Seichter, Daniel and Stephan, Benedict and Schmidt, Robin and Gross, Horst-Michael},
  booktitle={2025 IEEE/RSJ International Conference on Intelligent Robots and Systems},
  pages={2400--2407},
  year={2025},
  organization={IEEE}
}

\clearpage

\setcounter{page}{1}

\twocolumn[{%
\renewcommand\twocolumn[1][]{#1}%
\maketitlesupplementary

\begin{center}
\centering
\captionsetup{type=figure}
\vspace{-5mm}
\includegraphics[page=1, width=\textwidth]{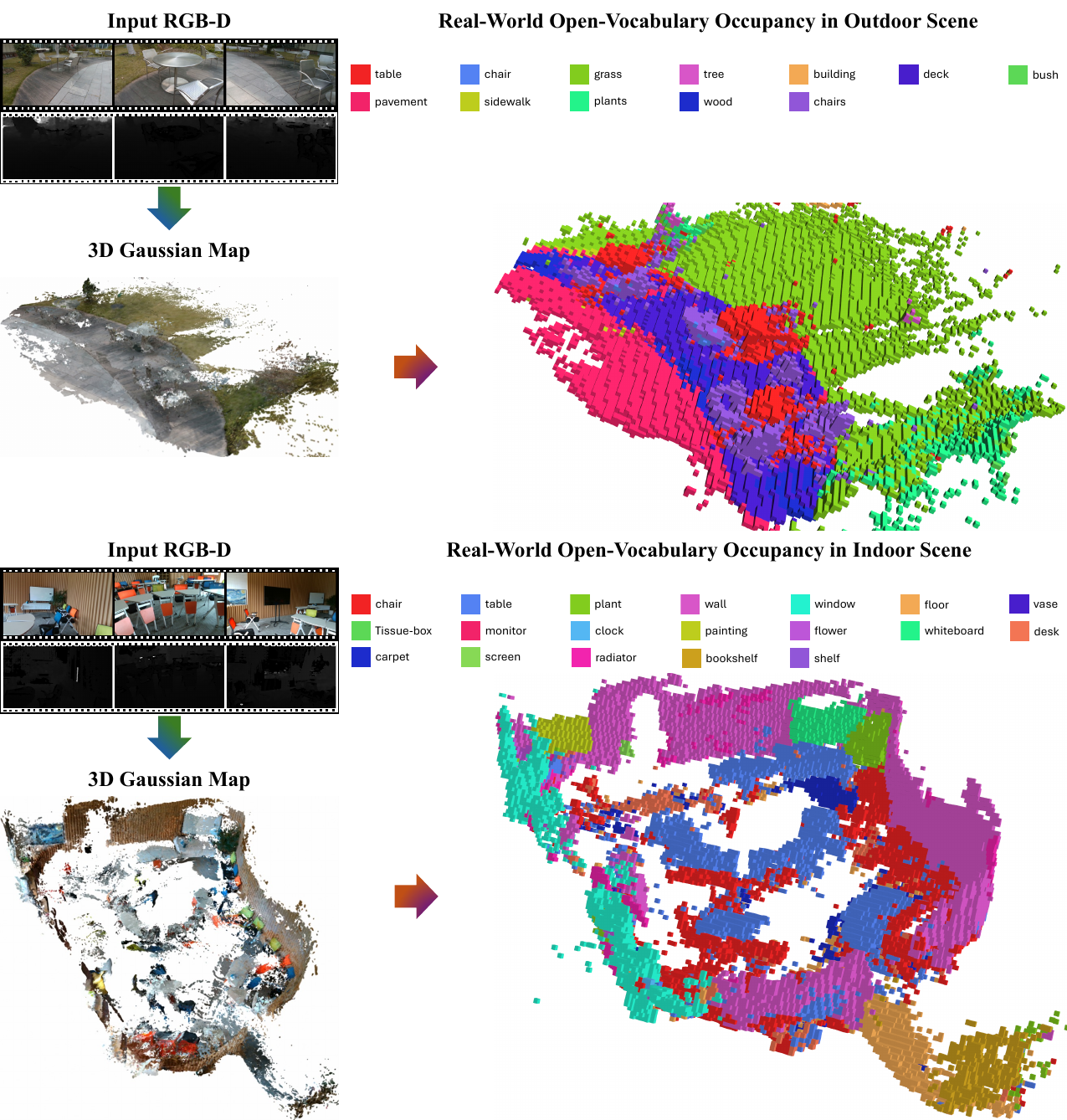} 
\vspace{-5mm}
\caption{This figure demonstrates the visualization results of \modelname{}'s open-vocabulary occupancy prediction in real-world indoor and outdoor scenes. For real-time results, please refer to the uploaded video.}
\label{fig: realworld}
\end{center}
}]

\section{Real-World Deployment with RGB-D Sensor}
\label{realworld}

In this section, we conduct open-vocabulary occupancy prediction experiments with \modelname{} in real-world environments. In contrast to benchmark experiments, real-world deployment provides neither pre-recorded trajectories nor ground-truth poses.
Such unconstrained sensing conditions directly reflect the target use case of \modelname{}, whose training-free, open-vocabulary formulation enables online occupancy construction from raw RGB-D streams without relying on pose supervision, closed-set labels, or offline optimization. As there are no other training-free baselines, we are the first to achieve a solution capable of deploying open-vocabulary occupancy prediction tasks in real-world settings.

All experiments in this paper were conducted on a device equipped with an Intel® Core™ i9-14900KF and a single NVIDIA GeForce RTX 5090.

\noindent\textbf{Online RGB-D acquisition.}
To validate the practicality of \modelname{} in real-world settings, we deploy the system with an Intel RealSense D435i RGB-D camera and run the full pipeline directly on live sensor streams.
During operation, the RGB and depth streams are synchronized and spatially aligned in real time, with depth measurements registered to the RGB camera frame.
The system selects the 1920$\times$1080 resolution for both color and depth streams, while maintaining real-time frame rates.
Depth values are acquired in raw sensor units and converted to metric scale using the device-reported depth factor $10^{-3}$.

\noindent\textbf{Open-vocabulary semantic cue generation.}
In real-world deployment, ground-truth semantic labels are not available.
To obtain open-vocabulary semantic cues from raw RGB observations, we leverage a pretrained Qwen3-VL vision-language model~\cite{bai2025qwen3} to generate all visible object categories for each incoming RGB frame.
During online mapping, we get scene-related labels of each predicted word across frames.
This temporal aggregation strategy yields a scene-level open-vocabulary semantic space without requiring any manual annotation or closed-set supervision.

The system takes the current RGB frame, the aligned depth map, and the semantic labels as input.
These inputs are directly fed into the \modelname{} pipeline, where camera poses are estimated incrementally, and a global 3D representation is updated online.
The incoming RGB-D observations are immediately integrated into the scene representation without requiring pre-recorded sequences or offline preprocessing.
To improve stability during real-world deployment, we apply a short warm-up period at the beginning of capture to allow the sensor auto-exposure to converge~\cite{teed2021droid, Homeyer_2025_ICCV}.
As shown in Fig.~\ref{fig: realworld}, this setup enables \modelname{} to be deployed in real-world environments as a streaming, open-vocabulary 3D occupancy-perception module for embodied agents.
\section{Exploratory Experiments}
\label{sec:exploratory}

In this section, we have included additional ablation experiments and sensitivity analyses of the system. This section includes three quantitative experiments and one qualitative experiment. These are intended to help future researchers better understand the system’s performance and potential avenues for further development.

\begin{table}[t]
\centering
\caption{
Influence of individual network components on EmbodiedOcc-ScanNet.
We evaluate different SLAM backbones and open-vocabulary semantic segmentation models under the monocular setting.
}
\label{tab:supp_component_ablation}
\setlength{\tabcolsep}{6pt}
\renewcommand{\arraystretch}{1.05}
\begin{tabular}{lccc}
\toprule
Method & IoU & mIoU & FPS \\
\midrule
\rowcolor{gray!15}
\multicolumn{4}{l}{\textbf{Default setting}} \\
Ours (mono) & 31.29 & 13.86 & 25.30 \\
\midrule
\rowcolor{gray!15}
\multicolumn{4}{l}{\textbf{SLAM backbone variants}} \\
MASt3R-SLAM & \textbf{33.80} & 15.66 & 18.10 \\
VGGT-SLAM   & 33.09 & \textbf{15.90} & \textbf{45.17} \\
\midrule
\rowcolor{gray!15}
\multicolumn{4}{l}{\textbf{VLM variants}} \\
SEEM        & 31.18 & 8.35  & 30.26 \\
DINOv2      & 31.59 & 8.18  & 24.93 \\
\bottomrule
\end{tabular}
\end{table}

\noindent\textbf{Influence of individual network components.}
To further evaluate the modularity of \modelname{}, we conduct additional ablation studies by replacing individual components while keeping the rest of the pipeline unchanged.
As shown in \cref{tab:supp_component_ablation}, replacing the default SLAM backbone with recent end-to-end SLAM methods, including MASt3R-SLAM~\cite{murai2024_mast3rslam} and VGGT-SLAM~\cite{maggio2025vggt-slam}, improves both IoU and mIoU under the monocular setting.
This indicates that the proposed framework can directly benefit from stronger geometric estimation without requiring changes to the occupancy prediction pipeline.
In particular, MASt3R-SLAM achieves the highest IoU, while VGGT-SLAM obtains the best mIoU and FPS among the evaluated SLAM variants.
We further evaluate alternative open-vocabulary semantic segmentation models, including SEEM~\cite{zou2023segment} and DINOv2~\cite{oquab2024dinov}.
Although these variants maintain comparable IoU, their mIoU is notably lower than the default setting.
This suggests that the geometric occupancy estimation remains stable, while semantic occupancy quality is more sensitive to the open-vocabulary semantic module.
Overall, these results demonstrate the flexibility and extensibility of our modular design: stronger SLAM backbones can improve geometric consistency, and the segmentation component can be replaced depending on the desired trade-off between semantic quality and runtime efficiency.

\begin{table}[t]
\centering
\caption{
Performance gap analysis under the RGB-D setting on EmbodiedOcc-ScanNet.
We evaluate the influence of camera pose accuracy and the type of semantic prediction.
}
\label{tab:supp_gap_analysis}
\setlength{\tabcolsep}{6pt}
\renewcommand{\arraystretch}{1.05}
\begin{tabular}{lcc}
\toprule
Setting & IoU & mIoU \\
\midrule
Ours & 34.40 & 15.84 \\
GT Pose & \textbf{45.06} & 21.34 \\
Closed-set & 34.39 & 20.42 \\
GT Pose + Closed-set & 45.03 & \textbf{27.39} \\
\bottomrule
\end{tabular}
\end{table}

\noindent\textbf{Performance gap analysis.}
To better understand the remaining gap between \modelname{} and fully supervised occupancy prediction methods, we analyze the effects of camera pose accuracy and semantic prediction quality under the RGB-D setting.
As shown in \cref{tab:supp_gap_analysis}, replacing estimated poses with ground-truth poses improves IoU from 34.40 to 45.06 and mIoU from 15.84 to 21.34, indicating that pose accuracy and geometric alignment still affect the final occupancy quality.
Meanwhile, the achieved IoU is already competitive with recent supervised methods, showing that \modelname{} can provide strong geometric occupancy estimation in a training-free manner.
We also replace the open-vocabulary semantic module with a closed-set segmentation model, DVEFormer~\cite{fischedick2025efficient}, whose 40-class predictions are manually mapped to the 11 occupancy categories of EmbodiedOcc-ScanNet.
This closed-set variant keeps a similar IoU but improves mIoU from 15.84 to 20.42, suggesting that semantic category assignment is more consistent with the fixed benchmark taxonomy.
Combining ground-truth poses with closed-set semantics further increases mIoU to 27.39, demonstrating that accurate poses and benchmark-aligned semantics are complementary.
However, the per-class IoU still lags behind fully supervised methods for categories such as sofa, furniture, and other objects, suggesting that semantic occupancy supervision remains important for maximizing mIoU on fixed-label benchmarks.
These results show that the current performance gap stems primarily from pose alignment and semantic category assignment, while also highlighting the potential of training-free, open-vocabulary occupancy prediction with stronger SLAM and VLM components.

\begin{table}[t]
\centering
\caption{
Quantitative open-vocabulary validation results on ReplicaOcc.
Categories are sorted by frequency from high to low, and mIoU is reported over the top-$K$ categories.
}
\label{tab:supp_open_vocab_replicaocc}
\setlength{\tabcolsep}{8pt}
\renewcommand{\arraystretch}{1.05}
\begin{tabular}{lcccc}
\toprule
Metric & Top-10 & Top-20 & Top-30 & Top-40 \\
\midrule
mIoU & 31.06 & 23.02 & 16.57 & 12.01 \\
\bottomrule
\end{tabular}
\end{table}

\noindent\textbf{Quantitative results in open-vocabulary validation.}
We further provide quantitative open-vocabulary validation results on ReplicaOcc.
We sort all categories by occurrence frequency and report mIoU over the top-$K$ categories, where $K$ is set to 10, 20, 30, and 40.
As shown in \cref{tab:supp_open_vocab_replicaocc}, \modelname{} achieves 31.06 mIoU on the top-10 categories and maintains 23.02, 16.57, and 12.01 mIoU when the vocabulary is expanded to the top-20, top-30, and top-40 categories, respectively.
The gradual decrease is expected, as lower-frequency categories often correspond to smaller objects, partial observations, or visually ambiguous regions, making them more challenging to detect without task-specific semantic occupancy supervision.

\begin{figure}[t]
\vspace{1mm}
\captionsetup{type=figure}
\includegraphics[width=\linewidth]{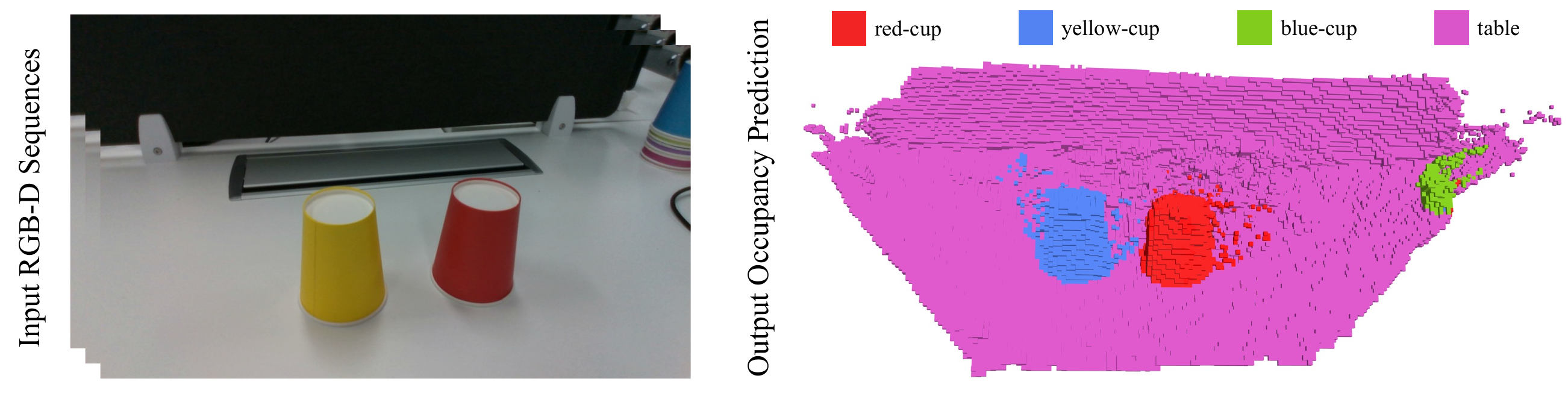}
\vspace{-2mm}
\caption{Real-world red-and-yellow cup experiment. \modelname{} correctly localizes and distinguishes visually similar objects according to open-vocabulary text queries, demonstrating its applicability to fine-grained real-world scene understanding.}
\label{fig:2cup}
\end{figure}

\noindent\textbf{Qualitative findings regarding desktop widgets.}
Since the standard grid size of 0.08 m is commonly used for individual-occupancy grids in embodied-occupancy prediction tasks, this choice significantly affects the accuracy of our reconstruction of small objects on a desk. As shown in Figure 2, we present the results of predicting the occupancy of paper cups of different colors on the table. However, we achieved better results only after reducing the side length of the individual occupancy grid to 0.005 m. This phenomenon also suggests that developing adaptive dynamic resolution for occupancy grids in embodied occupancy prediction is a highly promising area of research.

\begin{figure}[t]
\centering
\includegraphics[page=1, width=\linewidth]{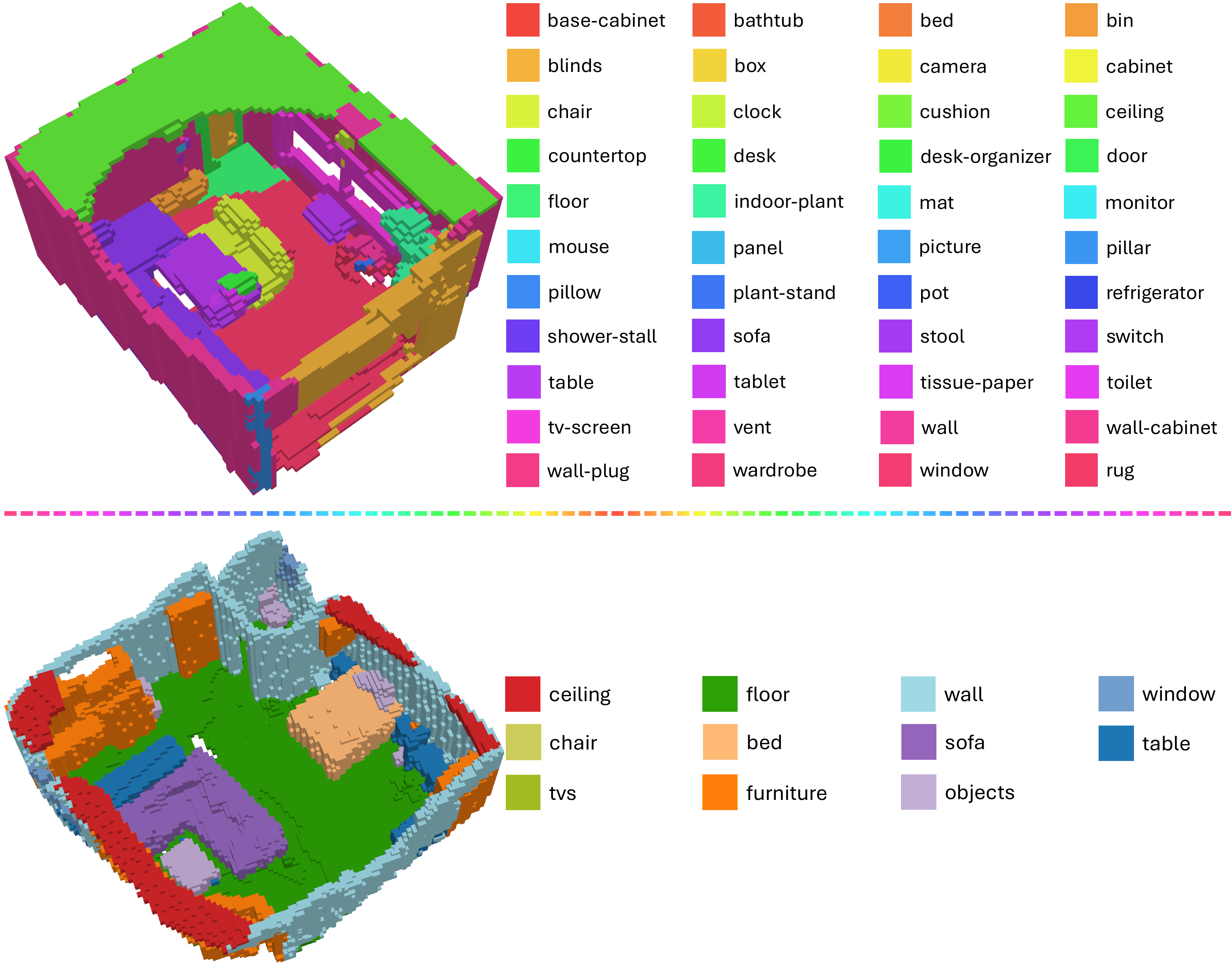} 
\caption{Visualization of a representative scene from EmbodiedOcc-ScanNet and ReplicaOcc with similar geometric layouts. While EmbodiedOcc-ScanNet contains 11 semantic categories, ReplicaOcc includes 44 categories per scene to maintain semantic diversity during evaluation.}
\label{fig: replicaocc}
\end{figure}

\begin{figure*}[t]
\centering
\includegraphics[page=1, width=\linewidth]{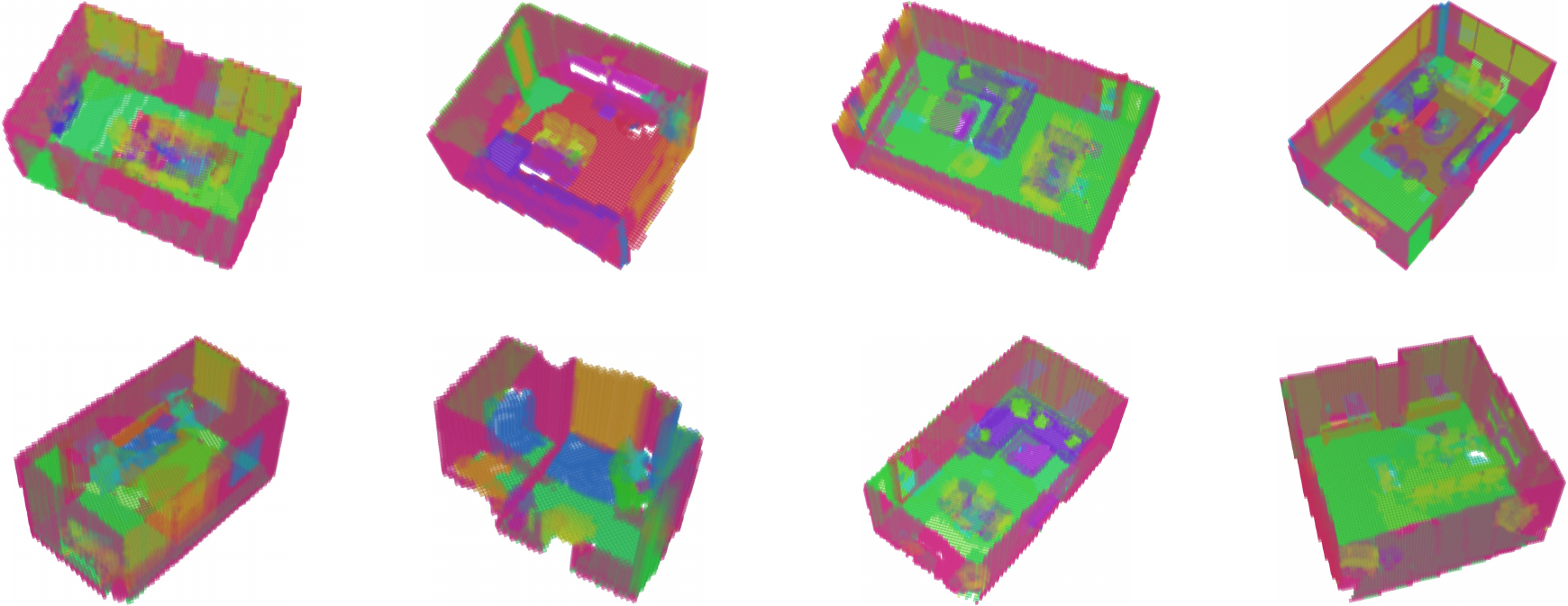} 
\caption{Visualization results for all ReplicaOcc scenes. In Replica, the true scene nearly touches the ceiling, so we reduced transparency to 0.3 for all ReplicaOcc visualization results.}
\label{fig: replicagt}
\end{figure*}

\begin{figure*}[t]
\centering
\includegraphics[page=1, width=\linewidth]{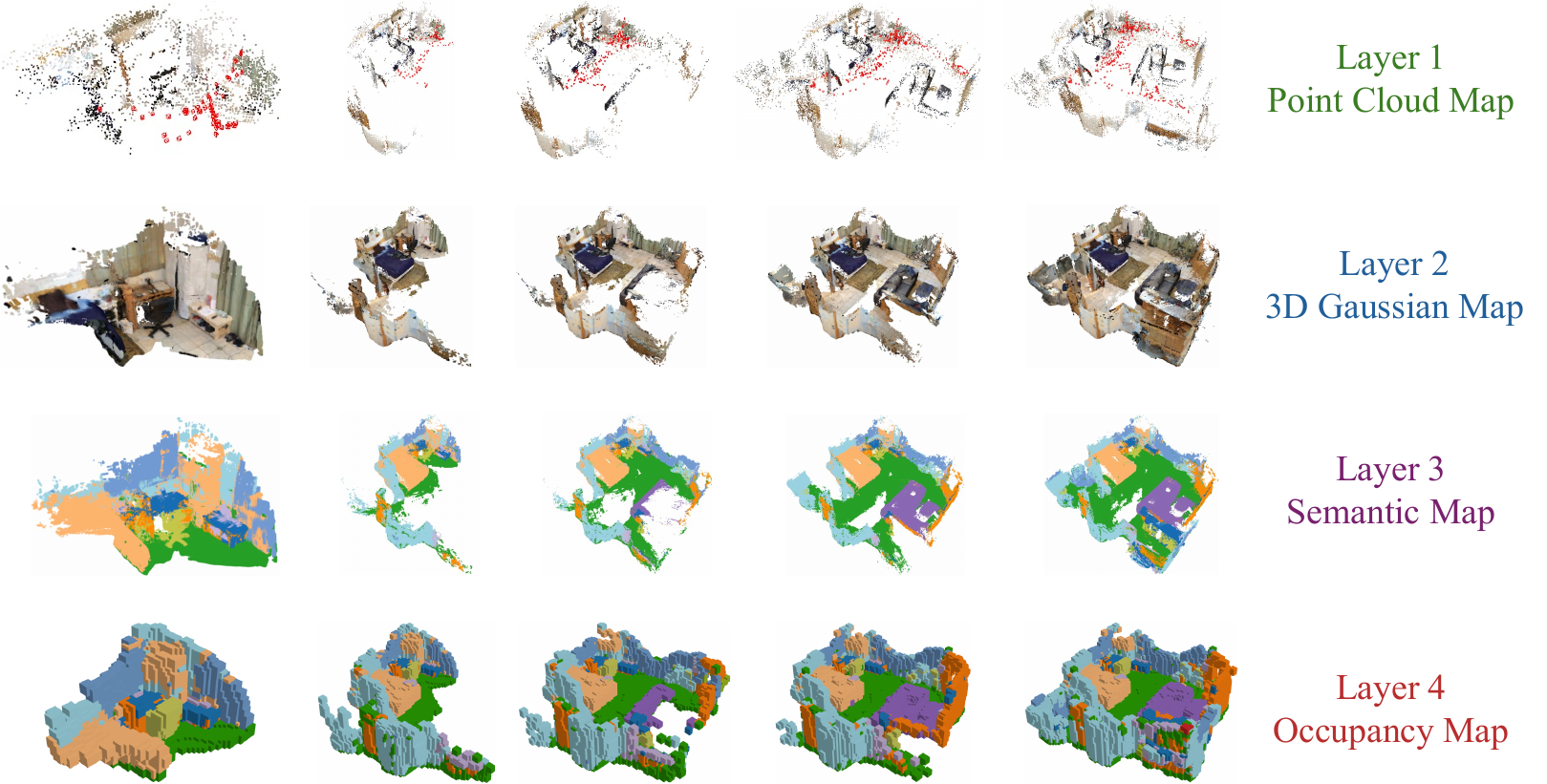} 
\caption{This figure displays the incremental results of multi-layer map construction for “scene0000” in ScanNet.}
\label{fig: incremental}
\end{figure*}

\begin{figure*}[t]
\centering
\includegraphics[page=1, width=\linewidth]{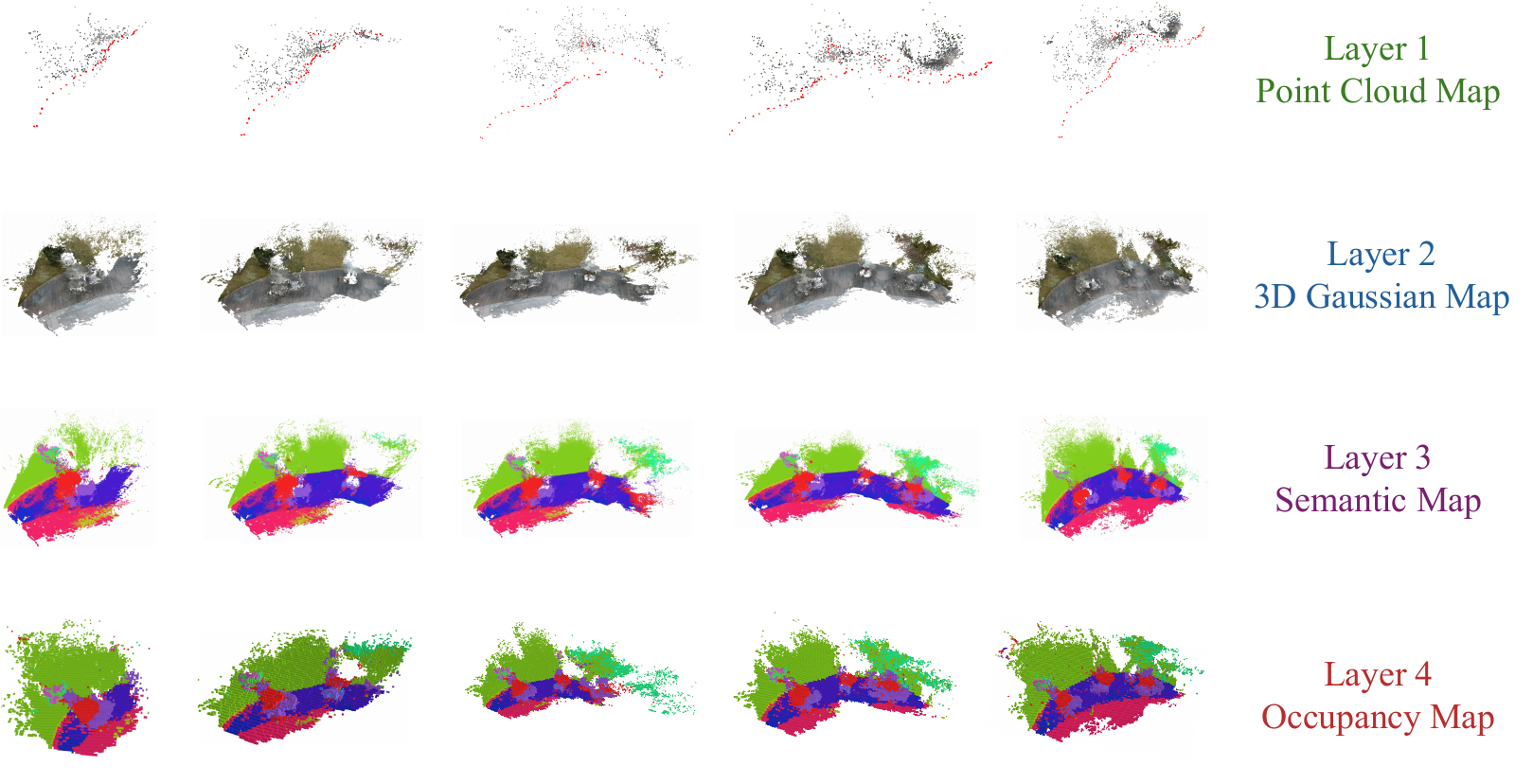} 
\caption{This figure displays the incremental results of multi-layer map construction for an outdoor scene in real-world deployment with RGB-D input.}
\label{fig: outdoor_incremental}
\end{figure*}

\section{Benchmark Details}
\label{sec:benchmark}

In this subsection, we describe the implementation details of the \textit{ReplicaOcc} benchmark.
Each Replica-SLAM scene~\cite{Zhu2022CVPR} provides RGB-D sequences, per-pixel semantic annotations, camera intrinsics, and camera poses.
Depth values are converted to metric scale using a scene-specific depth factor and truncated to a maximum range of $10\,\mathrm{m}$.
Following prior embodied occupancy benchmarks~\cite{ISO,embodiedocc}, we construct ReplicaOcc in three stages:
(i) extracting a sparse set of labeled voxels from RGB-D observations,
(ii) lifting them into a regular global voxel grid, and
(iii) determining voxel observability by fusing depth-frustum constraints over time.

\noindent\textbf{Sparse labeled voxel extraction from RGB-D.}
We first obtain a compact scene-level representation by back-projecting valid depth pixels into 3D space~\cite{ISO}.
Depth pixels are optionally subsampled with a fixed pixel stride $s_{\text{pix}}=4$ to reduce redundancy.
The resulting 3D points are transformed into the world frame using the corresponding camera poses and inherit semantic labels from the per-pixel annotations.
World points are quantized into voxels with a fixed voxel size $v=0.08\,\mathrm{m}$.
All points within each voxel are aggregated, and the voxel's semantic label is determined by majority voting over associated point labels.
This procedure yields a sparse set of labeled voxels that summarizes the observed scene geometry and semantics, which we store as scene-level preprocessed data.

\noindent\textbf{Construction of a regular global voxel grid.}
For evaluation, the sparse voxel set is further converted into a dense, regular grid covering the full spatial extent of the scene~\cite{ISO,embodiedocc}.
We align the sparse voxels by shifting them according to the scene-wise minimum coordinate and construct an axis-aligned voxel grid with the same resolution $v$.
The resulting grid has dimensions $N_x \times N_y \times N_z$, determined by the scene's spatial extent.
To populate the grid with semantic labels, each grid cell is assigned the label of its nearest sparse voxel if the distance is within 1 voxel; otherwise, it is initialized to an empty label.
This step produces a dense, globally consistent semantic grid that serves as the reference space for evaluating occupancy and semantic scene completion.

\noindent\textbf{Observability mask via fused depth-frustum consistency.}
Since not all voxels in the global grid are observable from the recorded camera trajectory, we explicitly compute a scene-level observability mask by fusing depth-frustum constraints~\cite{embodiedocc}.
A subset of frames is sampled along the trajectory with a frame stride $s_{\text{frm}}=2$.
For each selected frame, voxel centers are projected into the corresponding camera view using the known intrinsics and poses.
A voxel is considered observable if it lies in front of the camera, projects within the image boundaries, and is not occluded by the measured depth, allowing a tolerance proportional to the voxel size $v$.
The final observability mask is obtained by taking the union of observable voxels across all selected frames.
Optionally, a 3D binary dilation can be applied to mitigate discretization artifacts near surface boundaries.
Voxels outside this mask are treated as \emph{unknown} and excluded from evaluation (label 255), while observable but unlabeled voxels are regarded as known free space (label 0)~\cite{embodiedocc, gaussianformer}.

For each scene, we store the global voxel grid dimensions $(N_x,N_y,N_z)$, voxel center coordinates in the world frame, and dense semantic labels with unknown regions masked out, forming the final ReplicaOcc benchmark representation. Fig.~\ref{fig: replicaocc} illustrates representative scenes from EmbodiedOcc-ScanNet and ReplicaOcc. Fig.~\ref{fig: replicagt} illustrates the 8 scenes' visualization results of ReplicaOcc.

\section{Visualization Results}
\label{sec:visualization}

In this section, we present additional qualitative visualizations to illustrate the incremental mapping behavior of \modelname{}.
As shown in Fig.~\ref{fig: incremental}, our method progressively constructs a four-layer scene representation, including a point cloud map, a 3D Gaussian map, a semantic map, and a final occupancy map.
The point cloud layer provides sparse but reliable geometric anchors, while the 3D Gaussian layer densifies the observed regions and preserves richer surface appearance.
The semantic layer further associates language-aligned features with the reconstructed 3D structure, and the occupancy layer converts the accumulated geometric and semantic information into a voxelized representation for open-vocabulary occupancy prediction.

The visualization shows that the scene representation becomes increasingly complete as more frames are observed.
The semantic and occupancy maps remain well aligned with the underlying point cloud and Gaussian maps, suggesting that our open-vocabulary predictions are geometrically grounded rather than independently inferred from individual 2D frames.

Fig.~\ref{fig: outdoor_incremental} further presents real-world outdoor results.
Despite irregular geometry, larger depth variation, and more complex appearance, \modelname{} still constructs coherent multi-layer maps in an incremental manner.
The Gaussian layer provides a denser representation than the sparse point cloud, and the semantic and occupancy layers preserve meaningful region-level distinctions.

These qualitative results highlight the generalization ability of \modelname{}, showing that it can incrementally construct geometrically consistent and semantically meaningful open-vocabulary occupancy maps across both public benchmark datasets and real-world indoor/outdoor environments.

\end{document}